\documentclass[lettersize,journal]{IEEEtran}
\usepackage{amsmath,amsfonts}
\usepackage{algorithmic}
\usepackage{algorithm}
\usepackage{array}
\usepackage[caption=false,font=normalsize,labelfont=sf,textfont=sf]{subfig}
\usepackage{textcomp}
\usepackage{stfloats}
\usepackage{url}
\usepackage{verbatim}
\usepackage{graphicx}
\usepackage{multirow}
\usepackage{multicol}
\usepackage{colortbl}
\usepackage[normalem]{ulem}
\usepackage{thmtools}
\usepackage{amsmath, amssymb}
\usepackage{caption}
\usepackage{bm}
\usepackage{cite}
\usepackage{xcolor}
\newtheorem{theorem}{Theorem}

\definecolor{baselinecolor}{gray}{.9}
\newcommand{\baseline}[1]{\cellcolor{baselinecolor}{#1}}

\begin{document}

% \title{A Sample Article Using IEEEtran.cls\\ for IEEE Journals and Transactions}
% \title{\textbf{$A^3$}:Ambiguous Aberrations Captured via Misleading-Learning for Facial Forgery Semantic Refinement}
% \title{Fair Semantics Refinement via Misleading-Learning for Facial Forgery Detection}
% \title{Forgery Semantics Decoupling and Refinement via Misleading-Learning for Fair and Generalizable Detection of Deepfake}
%\title{Forgery Semantics Refinement via Misleading-Learning for Fair Detection of Deepfake}
% \title{Forgery Semantics Refinement via Misleading-Learning for Fair Deepfake Detection}
\title{Redundant Semantic Environment Filling via Misleading-Learning for Fair Deepfake Detection}

\author{Xinan He, Yue Zhou,
Shu Hu,~\IEEEmembership{Member,~IEEE,}
Bin Li,~\IEEEmembership{Senior Member,~IEEE,}
Jiwu Huang,~\IEEEmembership{Fellow,~IEEE,}
Feng Ding$^{*}$
        % <-this % stops a space

\IEEEcompsocitemizethanks{
\IEEEcompsocthanksitem Xinan He and Feng Ding are with the School of Software, Nanchang University, Nanchang, Jiangxi, 330031, China. e-mail:(shahur@email.ncu.edu.cn, fengding@ncu.edu.cn)
\IEEEcompsocthanksitem Yue Zhou and Bin Li are with the Guangdong Key Laboratory of Intelligent Information Processing and Shenzhen Key Laboratory of Media Security, Shenzhen University, Shenzhen, 518060, China. e-mail:(2450042008@email.szu.edu.cn, libin@szu.edu.cn)
\IEEEcompsocthanksitem Shu Hu is with the Department of Computer and Information Technology, Purdue University, IN, 46202, USA. e-mail:(hu968@purdue.edu)
\IEEEcompsocthanksitem Jiwu Huang is with Guangdong Laboratory of Machine Perception and Intelligent Computing, Faculty of Engineering, Shenzhen MSU-BIT University, Shenzhen 518116, China. e-mail:(jwhuang@smbu.edu.cn)
% \IEEEcompsocthanksitem Jiwu Huang is with Amazon Prime Video, Sunnyvale, CA, 94089, USA. e-mail:(chunhaol@amazon.com)
}
\thanks{$^{*}$ Feng Ding is the corresponding author.}
}
\markboth{Journal of \LaTeX\ Class Files,~Vol.~14, No.~8, August~2021}%
{Shell \MakeLowercase{\textit{et al.}}: A Sample Article Using IEEEtran.cls for IEEE Journals}


\maketitle

\begin{abstract}

Detecting falsified faces generated by Deepfake technology is essential for safeguarding trust in digital communication and protecting individuals. However, current detectors often suffer from a dual-overfitting: they become overly specialized in both specific forgery fingerprints and particular demographic attributes. Critically, most existing methods overlook the latter issue, which results in poor fairness: faces from certain demographic groups, such as different genders or ethnicities, are consequently more difficult to reliably detect. To address this challenge, we propose a novel strategy called misleading-learning, which populates the latent space with a multitude of redundant environments. By exposing the detector to a sufficiently rich and balanced variety of high-level information for demographic fairness, our approach mitigates demographic bias while maintaining a high detection performance level. We conduct extensive evaluations on fairness, intra-domain detection, cross-domain generalization, and robustness. Experimental results demonstrate that our framework achieves superior fairness and generalization compared to state-of-the-art approaches. 
\end{abstract}


\section{Introduction}

\IEEEPARstart{D}{eepfake} detection has advanced rapidly in recent years. A major challenge arises because Deepfake datasets often feature a monoculture of facial characteristics \cite{lin2024preserving} and exhibit distinctive fingerprints from different face-swap methods. This leads to the detector suffering from dual-overfitting: it tends to specialize in both specific forgery fingerprints and particular demographic attributes. However, the most state-of-the-art methods ignore the latter, focusing only on discovering more common forgery fingerprints \cite{he2025vlforgery, zhou2025breaking}, leading to predictions biased by attributes such as gender, ethnicity, age, and even faces with lighter skin tones \cite{hazirbas2021towards, xu2022comprehensive}. To ensure consistent detection performance across demographic groups, detectors must not only capture common forgery cues but also identify forgery traces shared across different demographic labels.

\begin{figure}[t]
  \centering
  \includegraphics[width=0.5\textwidth]{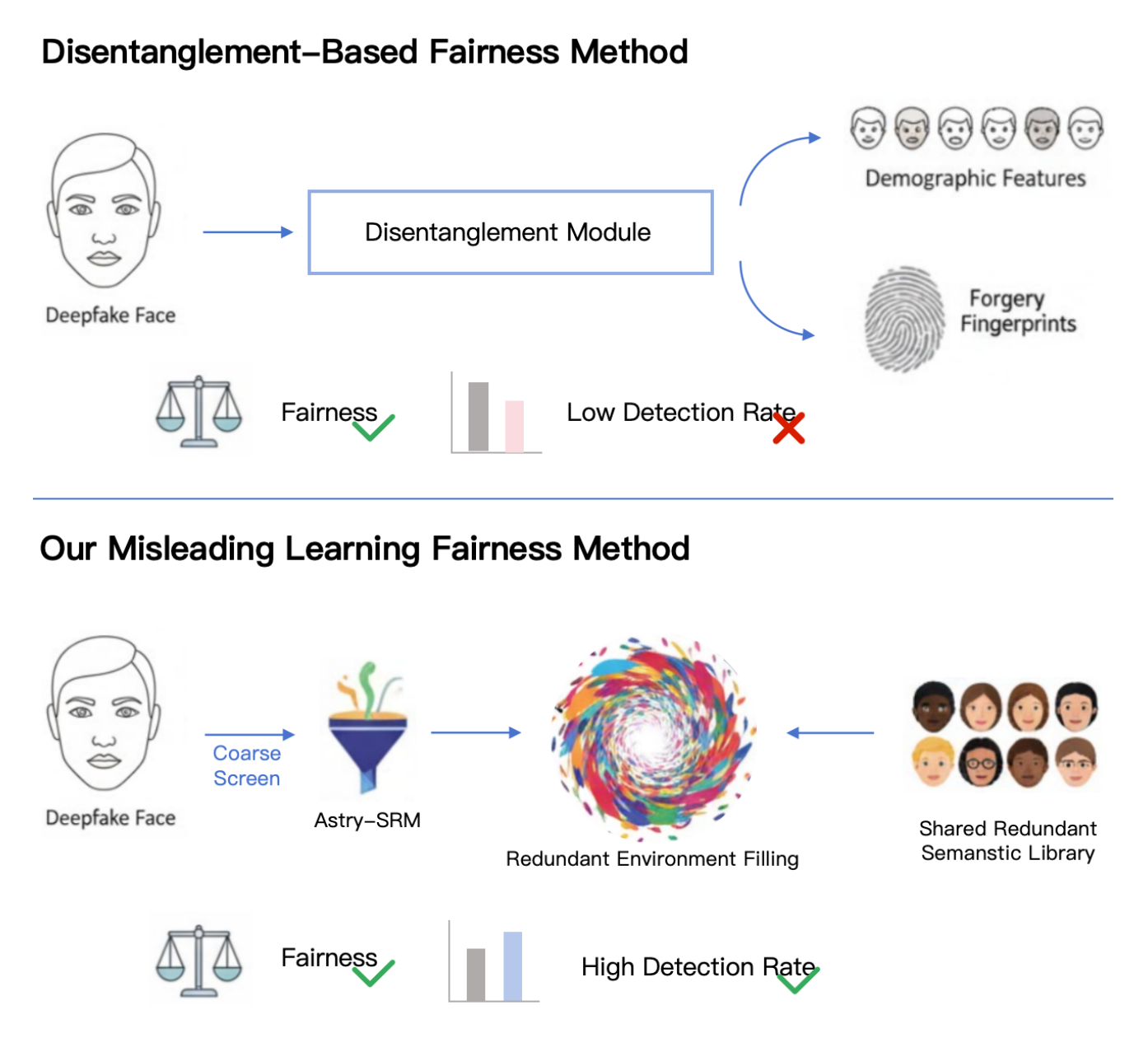}
  \caption{We compare our method with existing fairness approaches: {\em Disentanglement-based fairness method} aims at separating forgery fingerprints and demographic features through disentanglement learning; Our approach, termed {\em Misleading learning}, 
  enriches the redundant environment within the feature latent space.}
  \label{fig:int3}
  % \vspace{-10pt}
\end{figure}

Recently, some methods have also begun to acknowledge this problem and propose solutions. Ju \emph{et al.} \cite{Ju2024improving} proposed a method that can work with and without demographic annotations in the dataset
are based on distributionally robust optimization techniques and leverage specific loss functions to address the imbalance
in demographic groups and even in real and deepfake training samples.
However, this method performs well when training and testing data are generated by the same models but fails to ensure demographic fairness in cross-domain scenarios with unseen forgery generation methods. 
To improve demographic fairness generalization, Lin \emph{et al.} \cite{lin2024preserving} introduced a novel disentanglement framework
that separates demographic and domain-agnostic forgery features, thus enhancing the detector's fairness generalization. 

We contend that the primary cause of this demographic bias is that, within Deepfake datasets, the detector can't find absolutely pure forgery artifacts \cite{masood2023deepfakes}. Instead, the detector is compelled to integrate high-level semantic information to varying degrees as auxiliary cues for successful detection, leading to the phenomenon of dual-overfitting. Consequently, prior bias mitigation methods \cite{xu2022comprehensive, nadimpalli2022gbdf}, like disentanglement learning or fairness-aware loss functions, inadvertently cap the detector’s potential detection rate. This occurs because the bias mitigation process erroneously filters some of this performance-critical high-level semantic information.

However, exploring the creation of a training dataset that allows a detector to leverage high-level information for superior detection performance while simultaneously remaining nearly immune to the negative fitting effects (i.e., demographic bias) of that same high-level information remains a formidable challenge. This is because avoiding demographic bias requires the detector to be exposed to a sufficiently rich and balanced variety of high-level information.

To this end, we avoided the negative effects on detection rates commonly associated with disentanglement learning used in nearly all prior fairness methods. We instead propose a fundamentally different approach, which we term \emph{Misleading Learning}. The core idea is to disrupt the demographic distribution within the training data while simultaneously conducting various redundant environment filling within the latent space during feature extraction. This process is designed to simulate exposing the detector to a sufficiently rich and balanced variety of high-level information \cite{ding2021anti, ding2022securing,guan2022delving,chen2022ost}. Additionally, the method incorporates an \emph{Astry-SRM filter}, whose primary purpose is to perform a coarse screening of high-level semantic information to alleviate the training burden on the detector. Our experiments confirm that this rough information screening provides a tangible level of assistance \cite{Shiohara_2022_CVPR,li2022wavelet}. Fig.~\ref{fig:int3} illustrates the differences between ours and other disentanglement-based fairness methods.

Our main contributions can be summarized as follows: 

\begin{enumerate}

\item We propose a novel learning strategy, termed \emph{Mislead-learning}. 
By populating the latent space with a multitude of redundant environments, this strategy is designed to simulate exposing the detector to a sufficiently rich and balanced variety of high-level information for demographic fairness.

\item  We designed an advanced frequency-enhanced approach, \emph{Astray-SRM},
which can adaptively update within the redundant environment provided during the misleading-learning process, enabling it to perform a coarse screening of high-level semantic information to alleviate the training burden on the detector.

\item 
We evaluate the proposed method with exhaustive experiments. Compared to other fairness methods, our approach reduces the detector's sensitivity to specific demographic semantics, thereby mitigating demographic bias while maintaining a high detection performance level.
\end{enumerate}

The remainder of the paper is organized as follows. We briefly survey the background in the following section. After that, we theoretically analyze the redundant semantics filling process in Section \uppercase\expandafter{\romannumeral3}. The proposed method is described in Section  \uppercase\expandafter{\romannumeral4}. The experimental results are reported and discussed in Section \uppercase\expandafter{\romannumeral5}. Finally, we present our conclusions.

\section{Related Work}
% \subsection{Deepfake Generation}
% So far, most DeepFakes are generated using technologies such as VAE and GAN \cite{goodfellow2020generative}. These methods are exceptional in generating deepfake images of remarkable verisimilitude, thus effectively misleading the human visual perception \cite{fan2024synthesizing}. In the field of facial manipulation, the methodologies can be divided into several key subdivisions. The first category is the duplication of facial features, achievable via approaches such as FaceSwap \cite{FaceSwap2018} and deepfake replacements \cite{li2020celeb}. The existing face-swapping datasets of this approach are %UADFV\cite{8630787}, D-TIMIT\cite{Korshunov2018DeepFakesAN}, 
% FF++ \cite{Rossler_Cozzolino_Verdoliva_Riess_Thies_Niessner_2019}, Celeb-DF \cite{li2020celeb}, and DFDC \cite{Dolhansky2020TheDD}. 

% The other type is the creation of strikingly realistic artificial faces using GAN-based technology \cite{Karras_Laine_Aila_2019}. The typical datasets belonging to this category are iFake-FaceDB \cite{9133490}, DFFD \cite{Dang_2020_CVPR}, Faces \cite{face-generator}; The alteration of video streams, as exemplified by techniques like Face2Face \cite{Thies_Zollhofer_Stamminger_Theobalt_Niessner_2016}. 

% In recent years, with the booming of generative models, particularly diffusion models\cite{sohl2015deep, rombach2022high}, has exhibited remarkable performance in producing highly realistic Deepfakes\cite{guarnera2024level}. It presents new challenges for forensics efforts.

\subsection{Deepfake Detection}

To date, numerous studies have been conducted on Deepfake forensics. In the early stages, researchers focused on extracting biological features from faces to identify traces of Deepfake manipulations \cite{lyu2022deepfake}. Subsequently, signal-level artifacts in Deepfake videos proved more effective than biological features for uncovering Deepfake fingerprints \cite{wang2022deepfake}. Following the exploratory phase, it was discovered that deep neural network (DNN) models could also effectively detect Deepfake videos, after training with sufficient data \cite{guo2023exposing, li2023forensic}. Therefore, the majority of current detection methods are data-driven DNNs that operate as binary classifiers to distinguish frames from videos.  

In recent years, other than the direct data-driven manner, Deepfake forensics has evolved to study the forgery traces left by Deepfake technologies. There are methods exposing Deepfake by intuitive data augmentation. Shiohara \emph{et al.} \cite{Shiohara_2022_CVPR} proposed SBI that swaps identities within images of the same person, effectively simulating the characteristic features of real human face exchanges. The data augmented by SBI can significantly boost the performance for predicting Deepfake. Yan \emph{et al.}\cite{yan2024transcending} proposed a heuristic approach to carry out data augmentation on latent spatial features. There are other technologies that augment data via Discrete Cosine Transform and wavelet transform \cite{wolter2022wavelet,li2022wavelet}.

Furthermore, some researchers tried to disclose the unique forgery artifacts of Deepfake. Cao \emph{et al.} \cite{cao2022end} introduced a forgery detection framework that relies on re-configurable classification learning. Some disentanglement works \cite{Yan2023UCFUC,ye2024decoupling,lin2024preserving,zhu2024tmfd} improve detection efficiency by decoupling features unrelated to forgery. Utilizing capsule networks\cite{nguyen2019capsule} is another approach for identifying a range of Deepfake productions.

\subsection{Fairness in Deepfake Detection}

Previous works have identified and analyzed demographic fairness issues in Deepfake detection. For instance, \cite{trinh2021examination, hazirbas2021towards} highlighted biases within Deepfake datasets and detection models, revealing discrepancies in error rates across different subgroups. Pu \emph{et al.} \cite{pu2022fairness} evaluated the fairness performance of the MesiInception-4 Deepfake detector on FF++ and found discrepancies in detection performance across the two genders. Xu \emph{et al.} \cite{xu2022comprehensive} conducted comprehensive analyses of the bias present in existing Deepfake detectors, enriching the datasets by providing additional annotations. Lin \emph{et al.} \cite{lin2024ai} proposed a large AI-generated dataset based on demographics and conducted a fairness benchmark evaluation. Recently, Deepfake fairness learning has been advanced through innovative methods\cite{masood2023deepfakes}. Nadimpalli \emph{et al.} \cite{nadimpalli2022gbdf} specifically introduced a gender-balanced dataset to mitigate performance biases related to gender. Other approaches have proposed fairness learning strategies or frameworks aimed at reducing biases. For example, Ju \emph{et al.} \cite{Ju2024improving} employed a specialized loss function to address the unfairness of Deepfake detectors across different demographic groups. Lin \emph{et al.} \cite{lin2024preserving} addressed the unfairness in cross-domain testing by proposing a disentanglement framework. Ding \emph{et al.} \cite{ding2024fairadapter} proposed a fairness adapter based on visual foundation models (VFMs) to mitigate biases in detecting AI-generated images across different categories. However, in their pursuit of better demographic fairness, these methods often erroneously filter out high-level features that are actually useful to the detector during the debiasing process, thereby detrimentally impacting the detector's potential performance. This paper proposed a fundamentally different approach, conducting various redundant environment filling within the latent space during the feature extraction process to reduce the demographic biases within the Deepfake detector.

\vspace{5mm}
\section{Demographic Semantic Redundancy Analysis and Motivation}

\subsection{The Root Cause of Demographic Unfairness}
\label{sec:rootcause}

\begin{figure}[h]
  \centering
  \includegraphics[width=0.49\textwidth]{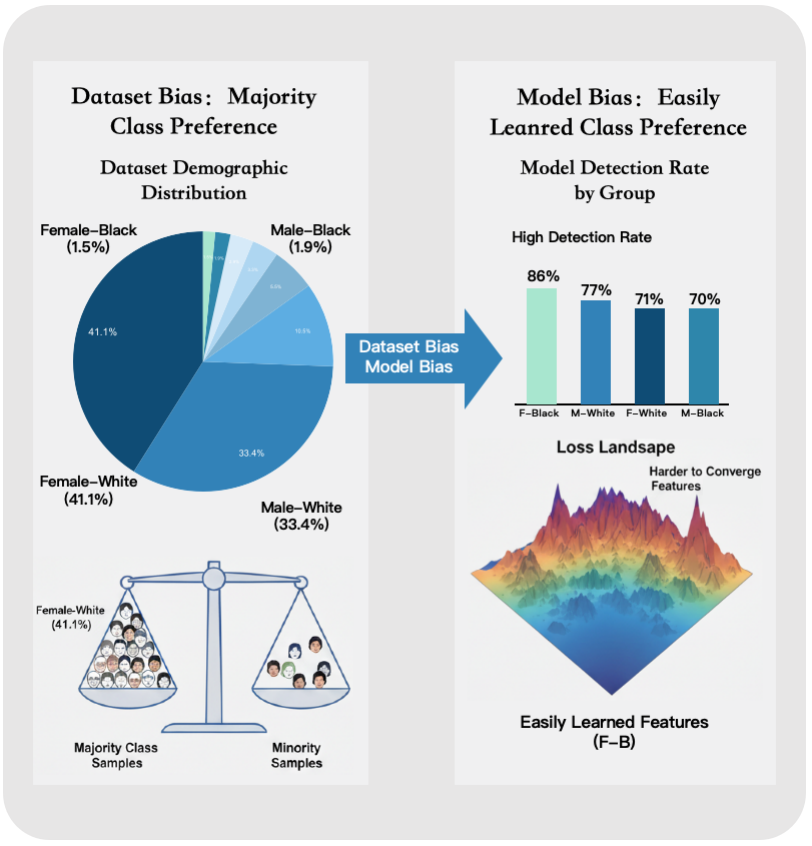}
  \caption{The left of the figure illustrates dataset bias, which stems from the imbalanced demographic distribution of the FF++ dataset. The lower right depicts model bias, resulting from the network's inherent architecture and initialization parameters. The upper right then shows how the combination of these two types of bias leads to the emergence of demographic unfairness.}
  \label{fig:pie}
  % \vspace{-10pt}
\end{figure}

We posit that a Deepfake face image fundamentally consists of pure forgery artifacts, irrelevant background, and facial feature information. Following image preprocessing (i.e., face cropping), the irrelevant background can effectively be disregarded in discussions of fairness and detection. The facial feature information itself is composed of characteristics like demographic attributes and other fine-grained features. Collectively, we refer to all features constituting the overall facial feature information as redundant features. The unfairness observed in detectors primarily manifests as a significant performance disparity in detecting face images with different redundant features. This disparity is particularly pronounced when dealing with highly distinctive redundant features such as gender and ethnicity.

We analyze the emergence of this unfairness from the following two angles  (see Fig.~\ref{fig:pie}):
\begin{enumerate}
    \item \textbf{Dataset Bias - Majority Class Preference}: In many large-scale benchmark datasets, the creators are often influenced by societal or cultural biases, inevitably resulting in a predominance of Caucasian faces among the image subjects. For instance, the demographic label distribution in the widely used FF++ dataset shows that Caucasian skin tones hold an absolute majority compared to other racial groups, and gender representation is also imbalanced. This skew causes the model during training to favor and fit the majority class samples, subsequently treating their redundant features as primary criteria for detection.
    \item \textbf{Model Bias - Easily Learned Class Preference}: Unfairness can also arise from the model itself due to Easily Learned Class Preference. The inherent initialization or architecture of the model's backbone network may create samples whose features occupy a region of the model's Loss Landscape that is easier to converge to.

\end{enumerate}

These two biases simultaneously cause the model to readily rely on redundant features to aid detection, resulting in the detector's dual-overfitting problem.

% \begin{figure*}[]
%   \centering
%   \includegraphics[width=\textwidth]{Figures/newfig2.pdf}
%   \caption{The whole process of misleading-learning. Deepfake can be understood as an entangled state of demographic semantics, forgery semantics, and other irrelevant semantics. The redundant semantics, which introduces ambiguity into the DeepFake detector, are depicted by the `$\times$' area within the circle labeled `forgery semantic'. To address this, our approach according to the demographics label of the Deepfakes, multiple negative samples with different demographics labels are randomly found in the negative sample bank to form the new negative sample set. The \textit{forgery artifacts extractor's}($i.e., f_e$) resistance to redundant semantics is enhanced through the extraction of forgery artifacts embedded within the hybrid feature set. As the training progresses, the extracted forgery semantics will become purer.}
%   \label{fig:representDis}
%   % \vspace{-10pt}
% \end{figure*}

\subsection{Limitations of Existing Fairness Decoupling Methods}

Existing fairness methods view the combination of forgery fingerprints and redundant features as an entangled feature representation. The use of these entangled features by current detectors often leads to the problem of unfairness. Specifically, we use variables: $X$ (e.g., an image), $Y$ (the target label of $X$, e.g., fake or real), $\hat{Y}$ (the classifier's prediction for $X$), $I$ (the redundant features, i.e., facial feature information). Then, we have the following Theorem.

\begin{theorem}
(\cite{locatello2019fairness}) If $X$ is entangled with $I$ and $Y$, the use of a perfect classifier for $\hat{Y}$, i.e., $P(\hat{Y}|X) = P(Y|X)$, does not imply demographic parity, i.e., $P(\hat{Y}=y|I=i_1)=P(\hat{Y} = y|I = i_2)$, $\forall y, i_1, i_2$.
\end{theorem}

Theorem 1 proposed by Locatello \textit{et al.}\cite{locatello2019fairness} emphasizes that different redundant features ($i.e.$,$i_1$, $i_2$) will induce specific interference in the result by directly using entangled semantics in a classifier. Most prior fairness methods assume that disentangling forgery fingerprints from redundant features is critical. Consequently, they employ strategies such as representation learning \cite{lin2024preserving} or utilize loss functions focused on demographic representation \cite{Ju2024improving} to mitigate the auxiliary influence of redundant features on the detector.

Although disentanglement learning has proven effective in other visual domains, we highlight two critical limitations in the context of Deepfake detection:

\begin{enumerate}
    \item \textbf{Difficulty in Isolating Fingerprints}: Forgery artifacts are inherently residual, incomplete, or inconsistent traces left by the generative model. They occupy a very small weight in the overall image feature space and are thus easily overwhelmed by high-level semantic information. Consequently, achieving a pure forgery artifact feature space through disentanglement operations amidst abundant high-level semantics is extremely challenging.

    \item \textbf{Loss of Useful Generalization Cues}: An overly aggressive removal of all redundant features risks the model losing features that are critical for generalization to new datasets or attack methods. For example, high-level features such as a significant difference in skin tone between the swapped face and the original identity, or abnormal face positioning, are themselves clear signs of unnatural forgery results. The detector can leverage these characteristics to identify inconsistencies.
\end{enumerate}

This is the central reason why these fairness methods fail to achieve a balance between demographic fairness and detection generalization capability.

% \begin{figure*}[tb]
%     \centering
%   \includegraphics[width = 1\textwidth]{Figures/Newfig4.pdf}
%   \caption{The overall overview of our proposed approach. \textbf{(1) } In the \textit{Prior Knowledge Acquisition} phase, we separately trained the astray detector and forgery artifacts extractor to obtain basic discrimination ability against forgery semantics. \textbf{(2) } 
%   % The \textit{Misleading Separation} phase is to decouple coarse and other redundant semantics from Deepfakes. The 
%   \textit{Misleading Separation} phase is to decouple demographic and other redundant semantics by enhancing latent space through negative samples with different races and genders. Simultaneously, it continuously searches for purer forgery semantics.
%   \textbf{(3) } In the \textit{Fine-tuning} phase, we use a new adapter to mitigate the information loss that occurred during the filtering of Astray-SRM.}
%   \label{fig:method}
%   % \vspace{-10pt}
% \end{figure*}

\begin{figure*}[tb]
    \centering
  \includegraphics[width = 1\textwidth]{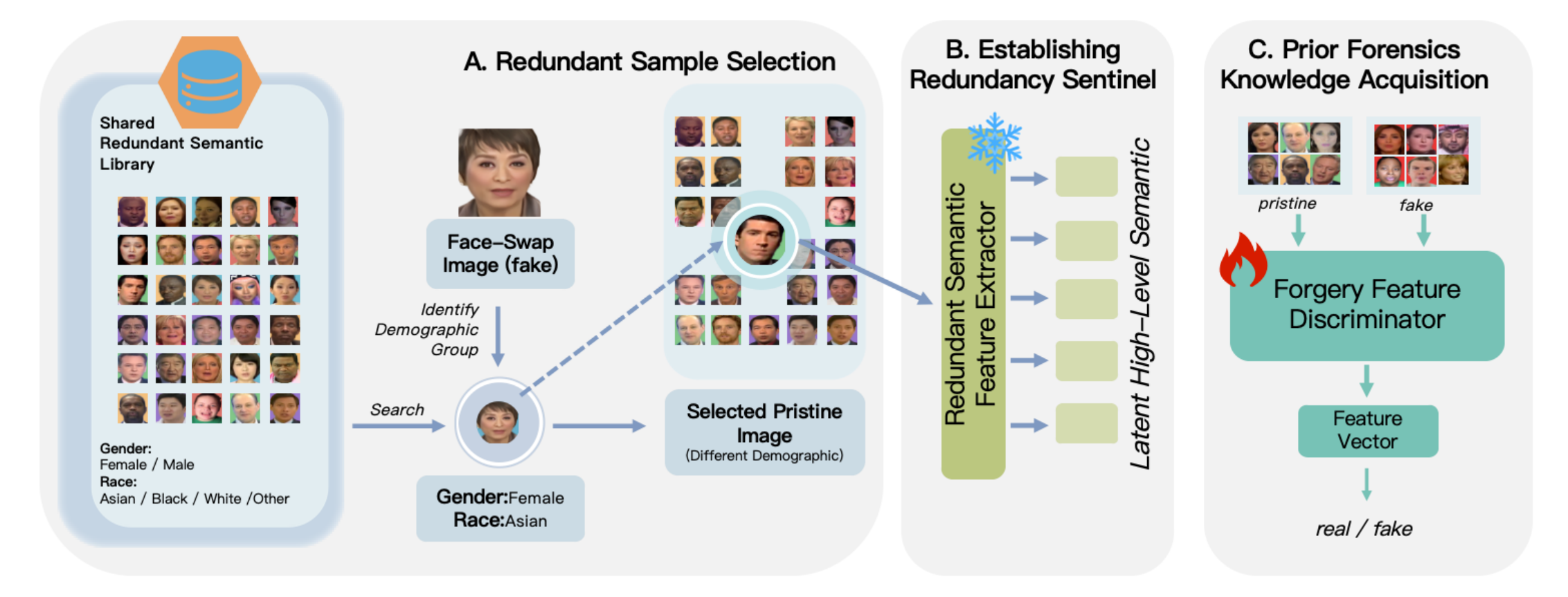}
  \caption{The first three stages of the Misleading-Learning methodology. (A) Redundant Sample Selection, where a redundant sample with differing demographic labels is chosen from the Shared Redundant Semantic Library based on the input fake image's group. (B) Establishing the Redundancy Sentinel, where the fixed and frozen Redundant Semantic Feature Extractor extracts high-level semantic features from redundant samples using zero-shot pretrained weights. (3) Prior Forensics Knowledge Acquisition, where the Forgery Feature Discriminator is independently trained as a standard binary classifier to acquire basic forgery feature extraction capability using the Binary Cross-Entropy loss.}
  \label{fig:method1}
  % \vspace{-10pt}
\end{figure*}

\subsection{Design Philosophy of Misleading Learning}
\label{sec:designpilosophy}

Rather than attempting to `filter out' redundant features from entangled features, as prior work does, we instead aim to `compel' the feature extractor to overcome the inherent biases of both the dataset and the model itself by creating an extremely rich and diverse redundant environment. This novel approach allows us to achieve fairness across samples with various redundant features without sacrificing the detector's intrinsic forensic capability.

\smallskip
\noindent
\textbf{Strategy to Address Model Bias.} First, to combat Model Bias caused by Easily Learned Class Preference, our core idea is to utilize the original identity image set (a superset of real images) as a shared library of redundant semantics. During training, whenever a fake image is input, we search this shared library for samples that are randomly diversified across identity, race, and gender. We then compel the fake sample's semantic features in the latent space to align with these diverse redundant samples. This achieves a multi-feature redundant environment (e.g., multi-skin tone) within the latent space, thereby weakening the detector’s reliance on any single redundant feature.

\smallskip
\noindent
\textbf{Strategy to Address Dataset Bias.} Regarding Dataset Bias caused by Majority Class Preference, we propose a randomized sampling strategy designed for these diversified redundant samples, detailed in the following formula:

\begin{equation}\small
    \mathcal{B}_{\mathcal{G}_g} := \frac{\frac{1}{\mathcal{D}_{\mathcal{G}_g}}}{\sum_{\mathcal{G}_i\in\mathcal{G}}(\frac{1}{\mathcal{D}_{\mathcal{G}_i}})},
\end{equation}

where $\mathcal{G}$ is the set of subgroups with each subgroup $\mathcal{G}_g \in \mathcal{G}$, $\mathcal{B}$ is the final selection bias for real faces across various demographic subgroups, $\mathcal{D}$ is the proportion threshold for various demographic subgroups within the original dataset.

\section{Misleading-Learning}
\label{sec:Deepfake Misleading Learning}

\begin{figure*}[tb]
    \centering
  \includegraphics[width = 1\textwidth]{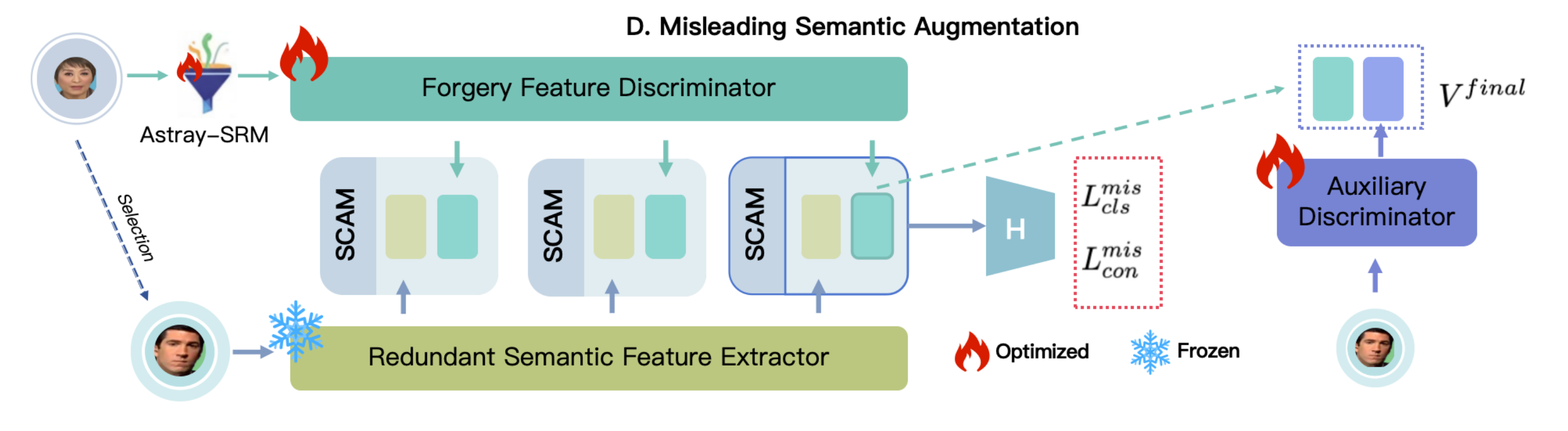}
  \caption{This is the pipeline for the fourth stage of misleading learning. It includes Astray-SRM for coarse-grained filtering of high-level semantic information, the construction of the Semantic Constraint and Augmentation Module, and the design of the Auxiliary Discriminator.}
  \label{fig:method2}
  % \vspace{-10pt}
\end{figure*}

Our proposed Misleading Learning methodology aims to resolve the generalization and demographic fairness issues in deepfake detection that stem from the strong entanglement between high-level semantic redundancy (such as identity and ethnicity) and forgery fingerprints. The core principle of the entire framework is to manually construct a rich redundant environment to compel the detector to maintain its original detection performance. The framework is organized into the following stages: 1. \emph{Redundant Sample Selection} (Acquiring the source material for the redundant environment), 2. \emph{Establishing the Redundancy Sentinel} (Creating a mapper from redundant samples to the redundant environment), 3. \emph{Prior Forensics Knowledge Acquisition} (Establishing the injection target for the redundant environment), 4. \emph{Misleading Semantic Augmentation} (Covering the specific injection procedure and the model's training strategy). The first three stages are detailed in Fig.~\ref{fig:method1}, and the fourth stage is presented in Fig.~\ref{fig:method2}.

\subsection{Redundant Sample Selection}

We maintain a Shared Redundant Semantic Library, $S_r=\{R_i\}^n_{i=1}$ with size $n$, where $R_i$ represents an available original identity sample, categorized by its demographic labels (gender and race). Given a face-swap (fake) image, $X_i$, we first identify its demographic group. To maximize semantic disparity, we must select an original image, $R_i$, from $S_r$ that belongs to a different demographic subgroup. We define the randomized sampling function $\mathcal{S}$ to execute this selection, detailed in the following formula:
\begin{equation}
    R_i = \mathcal{S}(X_i, S_r; \mathcal{B}_{\mathcal{G}_g}),
\end{equation}
where the probability distribution for sampling is driven by the bias metric $\mathcal{B}_{\mathcal{G}_g}$ (as defined in Section~\ref{sec:designpilosophy}).

\subsection{Establishing the Redundancy Sentinel}

In this stage, we introduce a core component: the Redundant Semantic Feature Extractor ($E^{red}$). This module's main function is to extract high-level semantic information that is independent of pure forgery artifacts (i.e., the redundant features discussed in Section~\ref{sec:rootcause}, such as gender and ethnicity). Specifically, it is used to extract the latent features from the samples selected from the Shared Redundant Semantic Library.

We implement this using a zero-shot approach, meaning $E^{red}$ requires no additional training. We load officially released pretrained weights (e.g., weights pretrained on ImageNet) for the chosen backbone network (such as Xception or ResNet), as these weights already possess strong visual representation capabilities. Furthermore, $E^{red}$ will operate as a fixed and frozen redundant feature extractor during the Misleading Semantic Augmentation stage. Because its weights were pretrained on a large-scale dataset and not fine-tuned on Deepfake data, it can efficiently extract the image's high-level semantic features with minimal attention paid to forgery fingerprints.

The feature extraction procedure for the $E^{red}$ is as follows:
\begin{equation}
    V^{red}_i = E^{red}(R_i)
\end{equation}
where $R_i$ presents the redundant sample selected based on the input fake image,  $V^{red}_i$ refers to the latent high-level semantic captured by $E^{red}$.

\subsection{Prior Forensics Knowledge Acquisition}
\label{sec:priorforens}

In this stage, we introduce another crucial component: the Forgery Feature Discriminator ($D^{sub}$). $D^{sub}$ is initially trained as an independent standard binary classifier to acquire basic forgery feature extraction capability. This is achieved by minimizing the Binary Cross-Entropy loss ($L_{bce}$) on the initial training set S, which grants it crude prior knowledge for distinguishing between real and fake images. Specifically, training set S = $\{(X_i, y_i)\}^n_{i=1}$ with size $n$. $X_i$ represents images (real or fake), $y_i$ represents the identity label $(y_i \in \{0:real, 1:fake\})$, indicating the source of $X_i$. The procedure is formulated as follows:
\begin{equation}
    L^{ini}_{cls} = L_{bce}(\mathcal{H}(D^{sub}(X_i)), y_{i}),
\end{equation}
where $\mathcal{H}$ denote the classify head for $D^{sub}$.

Subsequently, $D^{sub}$ 's parameters remain trainable throughout the following Misleading Semantic Augmentation stage. Its objective is to leverage the latent feature information extracted by itself (which contains forgery fingerprints) and the rich and diverse redundant environment provided by $E^{red}$. This is intended to guide the model to maintain its original detection performance, thereby reducing its reliance on specific redundant semantics and achieving demographic fairness.

\subsection{Misleading Semantic Augmentation}
The core objective of this stage is to determine how to effectively inject the constructed redundant environment into the latent features extracted by $D^{sub}$. (\textbf{Note}: We employ feature space fusion rather than direct pixel space replacement of image content, because directly blending the Deepfake image $X_i$ with the highly redundant original image $R_i$ in the pixel space would generate a large number of low-quality, semantically inconsistent samples that offer no assistance to detection and result in an excessive training load.)

\smallskip 
\noindent
\textbf{Astray-SRM in Misleading Semantic Augmentation.}

We discovered that directly fusing the raw latent features extracted by $D^{sub}$ (i.e., $V^{sub}$ ) presents the following issues:
\begin{enumerate}
    \item Although $D^{sub}$ undergoes basic training in Section~\ref{sec:priorforens}, the extracted $V^{sub}$ inherently carries a strong entanglement between forgery traces and high-level semantics (a common issue faced by all traditional Deepfake detectors). If we directly fuse this entangled $V^{sub}$ with $V^{red}$ in the feature domain, the model may tend to prioritize the easily learned high-level semantics within $V^{sub}$ (i.e., redundant information) as the key factor for feature fusion, rather than the pure forgery traces. This would strengthen $D^{sub}$'s reliance on redundant semantics, directly contradicting our objective.
    \item Without frequency-level processing, the fusion of $V^{sub}$ with $V^{red}$ would be dominated by high-level content in the feature dimensions (e.g., face shape, color). This fusion operation would dilute the subtle, original forgery fingerprints that were extracted, causing the model to overly focus on easily distinguishable semantic differences, which directly harms the model's core detection capability.
\end{enumerate}

Therefore, inspired by \cite{fei2022learning}, we designed the Astray-SRM in the proposed method ($i.e.$,$\delta_{astray-srm}$). We chose 3 kernels forming an array of 30, as they had been experimentally proven to be effective. Astray-SRM calculates the residual difference between the pixel value and the predicted values derived from its neighboring pixels. Details are depicted in the \textit{original} part of Fig~\ref{fig:astray-srm}.

And it's updated adaptively under the guidance of the Misleading Semantic Augmentation stage. The procedure is formulated as follows:
\begin{equation}
    % \delta_{astray-srm} = \delta_{srm} - \lambda(L_{misleading}/w)
    \delta_{astray-srm} = \delta_{srm} - \lambda \left(\frac{\partial L_{misleading}}{\partial w}\right),
\end{equation}
where $\delta_{srm}$ is the initial srm kernel before the update, $L_{misleading}$ is the loss function, specifically designed to penalize misleading-learning predictions (ref \ref{para:misleading_loss}: Misleading-Learning Loss). $\lambda$ is used to weight the regularization term in the loss function.
We conducted exhaustive experimentation, which encompassed the direct use of the original image, the comparison of results is shown in Sec~\ref{sec:Evaluation for Fair Generalizability of Different Image Pre-processes.}. 

In this stage, we input a pair of images: the deepfake image $X_i$ and its corresponding selected redundant sample $R_i$. $X_i$ will be pre-processed by the \emph{Astray-SRM} and then processed by the Forgery Feature Discriminator ($D^{sub}$) to extract the latent features $V^{sub}_i$. Concurrently, $R_i$ is processed by the Redundant Semantic Feature Extractor ($E^{red}$) to obtain the redundant environment features $V^{red}_i$. Subsequently, to efficiently and selectively utilize these features and complete the Misleading Semantic Augmentation, we need to perform both feature alignment and infusion operations on $V^{sub}_i$ and $V^{red}_i$. For this purpose, we introduce the \emph{SCAM module} (Semantic Constraint and Augmentation Module).
    
% \begin{figure}[t]

%   \includegraphics[width = 0.48\textwidth]{Figures/strurctures.pdf}
%   \caption{The architecture details of the SCAM block involves capturing forgery semantics from multiple dimensions, which are then fused with pristine image latent features processed through several convolutional blocks (labeled as `Conv Block' in the figure). }
%   \label{fig:scam-struct}
%   % \vspace{-10pt}
% \end{figure}

\smallskip
\noindent
\textbf{Semantic Constraint and Augmentation Module.} 
\label{para:Fusion Block of Single Channel Attention}

The SCAM module (Semantic Constraint and Augmentation Module) is designed to capture and enhance the latent features extracted by $D^{sub}$, which have been partially filtered of high-level information by the \emph{Astry-SRM}. Its goal is to enable effective alignment and fusion with the redundant features ($V^{red}$) extracted by $E^{red}$. The module contains multiple Alignment-Infusion Components ($F^{aic}$) that perform multi-semantic augmentation across several latent feature dimensions of both $D^{sub}$ and $E^{red}$, detailed in the following formula:

\begin{equation}
     V^{aug} = \mathcal{F}^{hy}(V^{red},\mathcal{F}^{aic}(V^{sub})),
\end{equation}
where $\mathcal{F}^{aic}$ denotes Alignment-Infusion Components block, $\mathcal{F}^{hy}$ denotes fuse block, $V^{aug}$ represents the latent feature after the redundant environment injection process. Specifically, $\mathcal{F}^{aic}$ first computes the input feature $V^{sub}\in \mathbb{R}^{h\times w\times c}$ to obtain an attention diagram $sc\in \mathbb{R}^{1\times 1\times c}$ for each channel. We calculate the mean and maximum values of samples from $V^{sub}$'s each channel by $\theta^{e}$ and $\theta^{a}$. $sc$ can be derived with the formula below:

\begin{equation}
    sc = sigmoid(conv(\theta^{e}(V^{sub}))+ conv(\theta^{a}(V^{sub}))).
\end{equation}
Note that each element in diagram $sc$ represents the purity each channel of $V^{sub}$ contains regarding forgery semantics. Subsequently, the regions that possess purer forgery semantics are enhanced by assigning larger weights to each channel, and vice versa:
\begin{equation}
    V^{sc} = V^{sub} + sc \otimes V^{sub},
\end{equation}
where $V^{sc}$ denotes the feature enhanced by $sc$. Then, we fuse the $V^{sc}$ and $V^{red}$ by $\mathcal{F}^{hy}$ module. The specific process is as follows:
\begin{equation}
    V^{aug} = \mathcal{F}^{hy}(V^{red}, V^{sc}) =  conv(concat(V^{red}, V^{sc})).
\end{equation}
The two features are concatenated, and the final mixed feature is obtained by a convolution block.

\begin{table*}[htbp]
\centering
\caption{\centering{Number of train/val/test samples of each subgroup in the FF++, Celebdf, DFD, and DFDC datasets. "-" means the group does not exist in the dataset.}}
\scalebox{1}{
\begin{tabular}{c|c|c|cccccccc}
\hline
\multirow{2}{*}{Dataset} & \multirow{2}{*}{-} &\multirow{2}{*}{Total Samples}&
\multicolumn{8}{c}{Intersection}\\ \cline{4-11}

&&&M-A& M-B&M-W&M-O&F-A&F-B&F-W&F-O \\ \hline

\multirow{3}{*}
{FF++}&Train&76,139&2,475&1,468&25,443&4,163&8,013&1,111&31,281&2,185\\ 
&Validation&25,386&2,085&480&8,308&1,409&1,428&150&10,627&899\\
&Test&25,401&1,167&602&9,117&1,171&2,182&598&9,723&841 \\ \hline

Celebdf&Test&28,458&-&1,922&924&60&-&-&25,432&120\\ \hline

DFD&Test&9,385&-&1,176&2,954&89&-&354&4,127&685 \\ \hline

DFDC&Test&22,857&548&2,515&6,460&1,648&658&3,385&6,441&1,202 \\ \hline
\end{tabular}
}
% \vspace{3mm}
\label{tab:the number of samples}
\end{table*}

\smallskip
\noindent
\textbf{Misleading Semantic Augmentation Loss.}

\label{para:misleading_loss}
Misleading Semantic Augmentation Loss is composed of two parts, Misleading Classification loss, and Misleading Contrastive loss.

\textit{\underline{Misleading Classification Loss.}} This loss penalizes classification errors made using $V^{aug}$. It compels the final latent feature ($V^{sub}_{final}$) of $D^{sub}$, after being injected with the redundant environment, to drive the classification head to classify the input as fake with high confidence. This process specifically serves to counteract the interference introduced by the redundant environment $V^{red}$.

Therefore, the mixed classification loss can be expressed as follows:

\begin{equation}
    L^{mis}_{cls}= L_{ce}(\mathcal{H}(V^{sub}_{final}), y_{f}),
    \label{eq:mixed ce}
\end{equation}

where $L_{ce}$ denotes the cross-entropy loss, $\mathcal{H}$ denote the classify head for $V^{sub}_{final}$. Note that $y_{f}$ is the classification label denoting $fake$.

\textit{\underline{Misleading Contrastive Loss. }}

We utilize a Regularization Contrastive Loss ($L_{con}$) to emphasize the disparity between $D^{sub}$'s final latent feature ($V^{sub}_{final}$) and $E^{red}$'s final latent feature ($V^{red}_{final}$). The primary function of this loss is to further constrain and mitigate the potential training difficulties that may arise from the overly complex injected redundant environment. The contrastive regulation loss is formulated mathematically as
\begin{equation}
    L^{mis}_{con} = max([m + \|f_{anchor}-f_+\|_2 - \|f_{anchor}-f_-\|_2] , 0),
    \label{eq:con}
\end{equation}
where $f_{anchor}$ denotes anchor features of an image, $f_+$ and $f_-$ denotes anchor features' positive samples and negative samples. $m$ denotes a hyper-parameter that quantifies the distances between the anchor, positive, and negative samples. if $f_+$ and $f_{anchor}$ denote the ($V^{sub}_{final}$), then $f_-$ represents ($V^{red}_{final}$) extracted by $E^{red}$.

\smallskip
\noindent
\textbf{Auxiliary Discriminator}

To ensure $D^{sub}$ maintains its fundamental discrimination ability on regular samples while learning fair forgery features, we introduce a lightweight auxiliary discriminator, $D^{aux}$. This also serves to compensate for the partial loss of high-level semantics caused by the Astry-SRM filter.

$D^{aux}$ is a structurally simplified feature extractor that receives the same input samples as $D^{sub}$ and extracts its own final latent feature ($V^{aux}_{final}$). This feature is then concatenated and fused with the feature ($V^{sub}_{final}$) extracted by $D^{sub}$ to form the final discriminative feature, $V_{final}$. The training objective of the entire $D^{sub}$ +$D^{aux}$  system also includes a Final Binary Classification Loss ($L^{final}_{cls}$ ) applied to the output derived from the fused feature, $V^{final}$, where y is the sample's ground truth label.
\begin{equation}
    L^{final}_{cls}= L_{ce}(\mathcal{H}(V_{final}), y_{i}),
    \label{eq:mixed ce}
\end{equation}
where $\mathcal{H}$ denote the classify head for $V_{final}$,  $y_i$ represents the identity label $(y_i \in \{0:real, 1:fake\})$.

\smallskip
\noindent
\textbf{Misleading-Learning Loss.}

The final Misleading-learning loss function is the total of the Misleading Classification Loss, Misleading Contrastive Loss and Final Binary Classification Loss:
\begin{equation}
    L_{misleading} = L^{mis}_{cls} + \alpha L^{mis}_{con} + \beta L^{final}_{cls},
\label{eq:over loss}
\end{equation}
where $\alpha$, $\beta$ are hyper-parameters for balancing the overall loss.

% \vspace{5mm}
\section{Experiments}
\label{sec:experiments}

In this section, we outline all pertinent configurations of the experiments such as the selection criteria for datasets, the metrics used for evaluation, and any additional parameters or considerations that influenced the results. Also, the proposed method is thoroughly evaluated from multiple aspects.

\subsection{Settings}
\label{sec:settings}

\smallskip
\noindent
\textbf{Datasets.} To validate the generalization capability of our proposed model, we carried out comprehensive testing across four extensively recognized large-scale benchmark datasets: FaceForensics++ (FF++) \cite{Rossler_Cozzolino_Verdoliva_Riess_Thies_Niessner_2019}, DeepFakeDetection (DFD) \cite{googledeepfakes2019}, Deepfake Detection Challenge (DFDC) \cite{Dolhansky2020TheDD}, and Celeb-DF \cite{li2020celeb}. Among them, FF++ serves as our main training dataset, which is composed of falsified facial images generated by five distinct algorithms, including Deepfakes (DF) \cite{DeepFakes2017}, Face2Face (F2F) \cite{thies2016face2face}, FaceSwap (FS) \cite{FaceSwap2018}, NerualTexture (NT) \cite{thies2019deferred} and FaceShifter (FST) \cite{li2019faceshifter}. Note that, during the experimental phase, among the three compressed versions of the FF++ (the higher the compression level of a dataset, the more challenging it becomes to detect forgery traces): raw, lightly compressed (HQ), and heavily compressed (LQ), \textit{we default to using the HQ version unless specified otherwise.} Following previous works \cite{lin2024preserving, Ju2024improving}, for evaluating the fairness attribute, we divide the dataset into a combination of demographic labels: Male-Asian (M-A), Male-White (M-W), Male-Black (M-B), Male-Others (M-O), Female-Asian (F-A), Female-White (F-W), Female-Black (F-B) and Female-Others (F-O). The number of training/validation/test samples within each subgroup for the four datasets is presented in Table~\ref{tab:the number of samples}.
% A comprehensive description of the dataset is provided in Appendix~\ref{appendix: Dataset Settings}.

\begin{table*}[]
\centering
\caption{\centering{Intra-domain sub-dataset evaluation on FF++. All methods are trained on FF++, and tested on its test sub-datasets separated by five forgeries. Two metrics are used for evaluation: $F_{MAG}$ and $AUC$.}}
\scalebox{0.95}{
\begin{tabular}{c|c|cccccccccc}
\hline
\multirow{3}{*}{Method}&\multirow{3}{*}{publication}& \multicolumn{10}{c}{Sub-Dataset}\\ \cline{3-12} 

&&\multicolumn{2}{c|}{DF} & \multicolumn{2}{c|}{F2F} & \multicolumn{2}{c|}{FS} & \multicolumn{2}{c|}{NT} & \multicolumn{2}{c}{FST} \\ \cline{3-12}
&&\multicolumn{1}{c|}{$F_{MAG}$↓} & \multicolumn{1}{c|}{$AUC$↑} &
\multicolumn{1}{c|}{$F_{MAG}$↓} & \multicolumn{1}{c|}{$AUC$↑} & 
\multicolumn{1}{c|}{$F_{MAG}$↓} & \multicolumn{1}{c|}{$AUC$↑} & 
\multicolumn{1}{c|}{$F_{MAG}$↓} & \multicolumn{1}{c|}{$AUC$↑} & 
\multicolumn{1}{c|}{$F_{MAG}$↓} & \multicolumn{1}{c}{$AUC$↑}  
\\ \hline

% Ori\cite{Rossler_Cozzolino_Verdoliva_Riess_Thies_Niessner_2019}&ICCV'19&3.520&98.22&\underline{2.471}&98.94&6.663&97.70&\underline{12.175}&\underline{96.12}&5.259&\underline{98.55}\\
DAW-FDD\cite{Ju2024improving}&WACV'24&3.97&97.41&3.85&97.21&5.77&96.46&\underline{12.78}&\underline{92.07}&\underline{5.34}&96.33\\
DAG-FDD\cite{Ju2024improving}&WACV'24&14.36&96.25&5.02&97.74&\textbf{2.42}&\underline{98.82}&18.39&84.08&14.14&95.99\\
% UCF\cite{Yan2023UCFUC}&ICCV'23&\textbf{1.424}&\textbf{99.26}&3.760&98.86&2.848&99.24&11.680&94.98&\textbf{1.568}&\textbf{99.15} \\
PG-FDD\cite{lin2024preserving}&CVPR'24&\underline{3.64}&\underline{98.01}&\underline{3.59}&\underline{97.98}&4.98&97.59&14.10&92.01&8.17&\underline{97.27}\\ \hline
\baseline{Ours}&-&\baseline{\textbf{1.26}}&\baseline{\textbf{99.39}}&\baseline{\textbf{1.29}}&\baseline{\textbf{99.41}}&\baseline{\underline{4.38}}&\baseline{\textbf{98.67}}&\baseline{\textbf{12.20}}&\baseline{\textbf{93.78}}&\baseline{\textbf{4.20}}&\baseline{\textbf{98.62}}\\ \hline
\end{tabular}
}
\label{tab:intra-domain}
\end{table*}
\vspace{4pt}
\smallskip
\noindent
\textbf{Baseline Methods.} We consider four widely-used CNN architectures to validate the effectiveness of high-purity forgery semantics extracted by misleading-learning as introduced in Section~\ref{sec:Deepfake Misleading Learning} (\textit{i.e.}, Xception \cite{Chollet_2017}, ResNet-34 \cite{He_Zhang_Ren_Sun_2016}, ResNet-50 \cite{He_Zhang_Ren_Sun_2016}, EfficientNet-b4 \cite{tan2019efficientnet}). The fairness and generalizability of the proposed method are examined by comparison with existing deepfake detectors, which are categorized as follows: naive detectors include Xception \cite{Chollet_2017}, EfficiNet-B4 \cite{tan2019efficientnet}, Meso4 \cite{afchar2018mesonet}, MesoIncep \cite{afchar2018mesonet}, frequency type detectors include F3Net \cite{qian2020thinking}, SRM \cite{luo2021generalizing}, SPSL \cite{Liu_Li_Zhou_Chen_He_Xue_Zhang_Yu_2021}, spatial type detectors include CORE \cite{ni2022core}, CapsuleNet \cite{nguyen2019capsule}, Dong \emph{et al.} \cite{nguyen2019capsule}, UCF \cite{Yan2023UCFUC}, fairness-enhanced detectors include DAW-FDD \cite{Ju2024improving}, DAG-FDD \cite{Ju2024improving}, PG-FDD \cite{lin2024preserving}.
\vspace{4pt}

\smallskip
\noindent
\textbf{Implementation Details.} All of our experiments are based on the PyTorch and trained with two NVIDIA RTX 4090Ti. For training, we use the Adam for optimization with the learning rate of $\beta=$\( 1 \times 10^{-3} \), and batch size of 16, for 20 epochs. For the overall loss, we set the $\alpha$ in Eq.~(\ref{eq:over loss}) as 0.05.
\vspace{4pt}

\smallskip
\noindent
\textbf{Data Preprocessing and Augmentation.} The input image size for all methods utilized in both training and testing is consistently set to 224x224 pixels. In order to enhance the model's robustness to post-processed images and ensure the fairness of the comparative experiments, a series of data augmentation techniques are applied to the images prior to their input into the model, including horizontal flipping, random rotation, Gaussian blur, random color adjustments, and image compression.

% For the mixed classification loss in Eq.~(\ref{eq:mixed ce}), we set the $\alpha_1$ as 1, for the overall loss, we set the $\alpha_2$ in Eq.~(\ref{eq:over loss}) as 0.05.

% and for the misleading contrastive loss, we set the $\alpha_2$ as 0.5. For the overall loss, we set the $\alpha_3$ in Eq.~(\ref{eq:over loss}) as 0.05. 
\vspace{4pt}

\smallskip
\noindent
\textbf{Evaluation Metrics.} To ascertain the efficacy of our proposed method, we adopt the Area Under the Curve (AUC) of the Receiver Operating Characteristic curve as our principal metric for evaluation, drawing upon the benchmarks established by preceding research.  Concurrently, we use three additional fairness metrics to provide a more comprehensive assessment. Specifically, we report the Maximum Area Under the Curve Gap ($F_{MAG}$), the Equal False Positive Rate ($F_{FPR}$), and Max Equalized Odds ($F_{MEO}$). 

% \begin{align*}
% \small
%     % &F_{FPR} := \sum_{\mathcal{G}_g\in\mathcal{G}} \left| \frac{\sum_{g=1}^{n} \mathbb{I}_{[\hat{Y}_g = 1, S_g =  \mathcal{G}_g, Y_g = 0]}}{\sum_{g=1}^{n} \mathbb{I}_{[S_g = \mathcal{G}_g, Y_g = 0]}} - \frac{\sum_{g=1}^{n} \mathbb{I}_{[\hat{Y}_g = 1, Y_g = 0]}}{\sum_{g=1}^{n} \mathbb{I}_{[Y_g = 0]}} \right|,\\
%     &F_{FPR} := \sum_{\mathcal{G}_g\in\mathcal{G}} \left| \frac{\sum_{g=1}^{n} \mathbb{I}_{[\hat{Y}_g = 1, S_g =  \mathcal{G}_g, Y_g = 0]}}{\sum_{g=1}^{n} \mathbb{I}_{[S_g = \mathcal{G}_g, Y_g = 0]}} \right.\quad\\
%      & \quad\quad\quad\quad\quad\quad\quad\quad \left. - \frac{\sum_{g=1}^{n} \mathbb{I}_{[\hat{Y}_g = 1, Y_g = 0]}}{\sum_{g=1}^{n} \mathbb{I}_{[Y_g = 0]}} \right|,\\
%     &F_{MAG} := \max_{G_g \in \mathcal{G}} \left\{ \frac{\sum_{g=1}^{n} \mathbb{I}_{[\hat{Y}_g=Y_g, S_g=G_g]}}{\sum_{g=1}^{n} \mathbb{I}_{[S_g=G_g]}} \right.\quad\\
%     & \quad\quad\quad\quad\quad\quad\quad\quad \left. - \min_{G_g' \in \mathcal{G}} \frac{\sum_{g=1}^{n} \mathbb{I}_{[\hat{Y}_g=Y_g, S_g=G_g']}}{\sum_{g=1}^{n} \mathbb{I}_{[S_g=G_g']}} \right\}, \\
%     &F_{MEO} := \max_{k, k' \in \{0,1\}} \left\{ \max_{G_g \in \mathcal{G}} \frac{\sum_{g=1}^{n} \mathbb{I}_{[\hat{Y}_g=k, Y_g=k', S_g=G_g]}}{\sum_{g=1}^{n} \mathbb{I}_{[S_g=G_g, Y_g=k]}} \right. \quad\\
%     & \quad\quad\quad\quad\quad\quad\quad\quad \left.- \min_{G_g' \in \mathcal{G}} \frac{\sum_{g=1}^{n} \mathbb{I}_{[\hat{Y}_g=k, Y_g=k', S_g=G_g']}}{\sum_{g=1}^{n} \mathbb{I}_{[S_g=G_g', Y_g=k]}}\right \}.   
% \end{align*}
\begin{equation}\small
    F_{FPR} := \sum_{\mathcal{G}_g\in\mathcal{G}} \left| \frac{\sum_{g=1}^{n} \mathbb{I}_{[\hat{Y}_g = 1, S_g =  \mathcal{G}_g, Y_g = 0]}}{\sum_{g=1}^{n} \mathbb{I}_{[S_g = \mathcal{G}_g, Y_g = 0]}} - \frac{\sum_{g=1}^{n} \mathbb{I}_{[\hat{Y}_g = 1, Y_g = 0]}}{\sum_{g=1}^{n} \mathbb{I}_{[Y_g = 0]}} \right|,
\end{equation}

\begin{equation}
\small
\begin{aligned}
    &F_{MAG} := \\
    &\max_{\mathcal{G}_g \in \mathcal{G}} \left\{ \frac{\sum_{g=1}^{n} \mathbb{I}_{[\hat{Y}_g=Y_g, S_g=\mathcal{G}_g]}}{\sum_{g=1}^{n} \mathbb{I}_{[S_g=\mathcal{G}_g]}}  - \min_{\mathcal{G}_g' \in \mathcal{G}} \frac{\sum_{g=1}^{n} \mathbb{I}_{[\hat{Y}_g=Y_g, S_g=\mathcal{G}_g']}}{\sum_{g=1}^{n} \mathbb{I}_{[S_g=\mathcal{G}_g']}} \right\}, 
\end{aligned}
\end{equation}

\begin{align}
\small
    F_{MEO} := \max_{k, k' \in \{0,1\}} \left\{ \max_{\mathcal{G}_g \in \mathcal{G}} \frac{\sum_{g=1}^{n} \mathbb{I}_{[\hat{Y}_g=k, Y_g=k', S_g=\mathcal{G}_g]}}{\sum_{g=1}^{n} \mathbb{I}_{[S_g=\mathcal{G}_g, Y_g=k]}} \right. \quad\nonumber\\
     \quad\quad\quad\quad\quad\quad \left.- \min_{\mathcal{G}_g' \in \mathcal{G}} \frac{\sum_{g=1}^{n} \mathbb{I}_{[\hat{Y}_g=k, Y_g=k', S_g=\mathcal{G}_g']}}{\sum_{g=1}^{n} \mathbb{I}_{[S_g=\mathcal{G}_g', Y_g=k]}}\right \}.   
\end{align}

Where $S$ is the demographic variable, $\mathcal{G}$ is the set of subgroups with each subgroup $\mathcal{G}_g'$, $\mathcal{G}_g\in\mathcal{G}$. $F_{FPR}$ measures the disparity in False Positive Rate (FPR) across different groups compared to the overall population. $F_{MAG}$ measures the maximum difference in the area under the curve across all demographic groups, and $F_{MEO}$ captures the largest disparity in prediction outcomes (either positive or negative) when comparing different demographic groups. 
    
The experimental results presented in all tables are used ↑ to indicate that higher values are better, and ↓ that lower values are better. By default, the best results are shown in \textbf{bold}, and the second-best results are indicated with \underline{underline}.

\begin{table*}[]  
\centering
\caption{\centering{Cross-domain dataset evaluation on $F_{MAG}$, $AUC$ metrics. All methods are trained on FF++ and evaluated on FF++, Celebdf, DFD, DFDC datasets. Top 3 values on each metric are highlighted in \colorbox[HTML]{9AFF99}{green,} \colorbox[HTML]{96FFFB}{blue,} \colorbox[HTML]{FFFC9E}{yellow.}}}
\scalebox{0.9}{
\begin{tabular}{c|c|c|ccccccccccc}
\hline
\multirow{2}{*}{\textbf{Method Type}}&\multirow{2}{*}{\textbf{Method}}&\multirow{2}{*}{\textbf{Backbone}}&\multicolumn{2}{c}{\textbf{FF++}}&\multicolumn{2}{c}{\textbf{Celebdf}}&\multicolumn{2}{c}{\textbf{DFD}}&\multicolumn{2}{c}{\textbf{DFDC}}\\ \cline{4-11}

&&&\multicolumn{1}{c|}{$F_{MAG}$↓} & \multicolumn{1}{c|}{$AUC$↑}&\multicolumn{1}{c|}{$F_{MAG}$↓} & \multicolumn{1}{c|}{$AUC$↑}&\multicolumn{1}{c|}{$F_{MAG}$↓} & \multicolumn{1}{c|}{$AUC$↑} & \multicolumn{1}{c|}{$F_{MAG}$↓} & \multicolumn{1}{c}{$AUC$↑}\\ \hline
% xce6 spsl6 core8 capsulenet12 Meso47 MesoIncep16 Dongeff7 srm6

% Xception\cite{Rossler_Cozzolino_Verdoliva_Riess_Thies_Niessner_2019}&Xception&4.82&97.43&17.77&77.82&12.74&77.91&34.90&62.07\\
\multirow{4}{*}{Naive}
% resnet32\\
% &resnet50\\
&Xception \cite{Chollet_2017}&Xception&5.82&\cellcolor[HTML]{96FFFB}97.43&17.72&77.82&\cellcolor[HTML]{FFFC9E}\cellcolor[HTML]{FFFC9E}12.74&77.91&34.90&62.08\\
&Efficient-B4 \cite{tan2019efficientnet}&Efficient-B4&8.79&93.87&22.19&77.62&30.28&78.80&38.29&62.60\\
&Meso4 \cite{afchar2018mesonet}&Meso4&11.82&69.08&17.72&65.70&16.90&56.51&62.77&58.42\\
&MesoIncep \cite{afchar2018mesonet}&MesoIncep&12.69&80.27&24.29&78.31&23.84&68.39&32.11&56.91\\ \hline
\multirow{3}{*}{Frequency}&F3Net\cite{qian2020thinking}&Xception&7.06&94.57&15.43&\cellcolor[HTML]{96FFFB}83.82&\cellcolor[HTML]{9AFF99}10.36&78.31&33.09&62.80\\
&SRM\cite{luo2021generalizing}&Xception&6.56&95.08&20.92&84.48&16.61&75.61&44.49&\cellcolor[HTML]{9AFF99}64.23\\
&SPSL\cite{Liu_Li_Zhou_Chen_He_Xue_Zhang_Yu_2021}&Xception&11.69&93.48&12.65&80.95&15.12&76.21&40.66&61.06\\ \hline
\multirow{5}{*}{Spatial}&CORE \cite{ni2022core}&Xception&9.85&94.32&22.99&79.70&13.61&76.72&40.50&61.92\\
&CapsuleNet \cite{nguyen2019capsule}&CapsuleNet&12.72&90.56&20.00&79.14&21.47&72.01&37.83&56.57\\
% &Recce\cite{cao2022end}&Designed&69.65&79.73&57.84&69.07\\
% Face X-ray\cite{li2020face}&CVPR'20&HRNet&&&\\
% SBI\cite{Shiohara_2022_CVPR} &EfficientNet&\underline{84.94}&78.93&62.21&\underline{75.36}\\
&Dong \textit{et al.}\cite{dong2023implicit}&EfficientNet&10.66&93.74&25.73&79.13&39.14&77.94&33.41&\cellcolor[HTML]{FFFC9E}62.93\\
&UCF\cite{Yan2023UCFUC}&Xception&5.86&\cellcolor[HTML]{FFFC9E}96.53&15.62&\cellcolor[HTML]{FFFC9E}82.82&20.74&\cellcolor[HTML]{96FFFB}82.74&43.72&61.76 \\ \hline
% TALL-Swin\\
% LSDA\\
\multirow{4}{*}{Fairness-enhanced}&DAW-FDD\cite{Ju2024improving}&Xception&\cellcolor[HTML]{FFFC9E}5.61&95.82&\cellcolor[HTML]{9AFF99}9.80&82.72&14.88&78.39&37.92&\cellcolor[HTML]{96FFFB}64.15\\
&DAG-FDD\cite{Ju2024improving}&Xception&11.68&94.35&18.61&81.39&15.80&80.56&\cellcolor[HTML]{96FFFB}31.01&56.84\\
&PG-FDD\cite{lin2024preserving}&Xception&\cellcolor[HTML]{96FFFB}5.16&96.31&\cellcolor[HTML]{FFFC9E}11.93&82.33&\cellcolor[HTML]{96FFFB}11.03&\cellcolor[HTML]{FFFC9E}81.53&\cellcolor[HTML]{9AFF99}27.76&62.13\\ \cline{2-11}

&Ours&Xception&\cellcolor[HTML]{9AFF99}4.90&\cellcolor[HTML]{9AFF99}97.89&\cellcolor[HTML]{96FFFB}10.12&\cellcolor[HTML]{9AFF99}85.11&13.99&\cellcolor[HTML]{9AFF99}86.70&\cellcolor[HTML]{FFFC9E}31.53&62.13\\ \hline
% \cellcolor[HTML]{9AFF99}  \cellcolor[HTML]{96FFFB}  \cellcolor[HTML]{FFFC9E}
\end{tabular}
}
\label{tab:cross-domain}
\end{table*}

\subsection{Evaluation for Fair Detection Performance on the Intra-domain Dataset}

\begin{table*}[t]
\centering
\caption{\centering{Fairness adaptability of different backbones. The evaluation encompasses $AUC$, $ACC$ metrics, and the $F_{FPR}$ and $F_{MEO}$ fairness metrics, across both intra-domain(FF++) and cross-domain(Celebdf, DFD, DFDC) datasets.}}
\scalebox{0.78}{
\begin{tabular}{c|c|cccc|cccc|cccc|cccc}
\hline
\multirow{3}{*}{Dataset} & \multirow{3}{*}{Method} &
\multicolumn{4}{c|}{Xception}& \multicolumn{4}{c|}{Resnet-50}&
\multicolumn{4}{c|}{Resnet-34}& \multicolumn{4}{c}{Efficientnet-b4}\\ \cline{3-18}

&&
\multicolumn{2}{c|}{\begin{tabular}[c]{@{}c@{}}Fairness\\ Metrics(\%)\textdownarrow\end{tabular}} & \multicolumn{2}{l|}{\begin{tabular}[c]{@{}c@{}}Detection\\ Metrics(\%)\textuparrow\end{tabular}} &
\multicolumn{2}{c|}{\begin{tabular}[c]{@{}c@{}}Fairness\\ Metrics(\%)\textdownarrow\end{tabular}} & \multicolumn{2}{l|}{\begin{tabular}[c]{@{}c@{}}Detection\\ Metrics(\%)\textuparrow\end{tabular}} &
\multicolumn{2}{c|}{\begin{tabular}[c]{@{}c@{}}Fairness\\ Metrics(\%)\textdownarrow\end{tabular}} & \multicolumn{2}{l|}{\begin{tabular}[c]{@{}c@{}}Detection\\ Metrics(\%)\textuparrow\end{tabular}} &
\multicolumn{2}{c|}{\begin{tabular}[c]{@{}c@{}}Fairness\\ Metrics(\%)\textdownarrow\end{tabular}} & \multicolumn{2}{l}{\begin{tabular}[c]{@{}c@{}}Detection\\ Metrics(\%)\textuparrow\end{tabular}}\\ \cline{3-18}

&&
$F_{FPR}$& \multicolumn{1}{c|}{$F_{MEO}$}& $AUC$ & $ACC$ & 
$F_{FPR}$& \multicolumn{1}{c|}{$F_{MEO}$} & $AUC$ & $ACC$  & 
$F_{FPR}$ & \multicolumn{1}{c|}{$F_{MEO}$} & $AUC$ & $ACC$ &
$F_{FPR}$ & \multicolumn{1}{c|}{$F_{MEO}$} & $AUC$ & $ACC$ \\ \hline
% ori: xce6 res507 res349 efficb48
% daw: xce9 res509 res349 efficb49
% dag: xce6 res508 res343
% pg: xce3 res507 res349 efficb44
% ours: xce18 res507 res348
\multirow{5}{*}{FF++}
&Ori~\cite{Rossler_Cozzolino_Verdoliva_Riess_Thies_Niessner_2019}&\underline{16.04}&\textbf{16.04}&\underline{97.43}&\textbf{92.98}&\underline{21.97}&\underline{21.97}&95.79&90.14&25.49&25.49&96.15&\textbf{92.81}&42.91&42.91&93.87&89.85\\
&DAW-FDD~\cite{Ju2024improving} &27.05&27.05&95.82&\underline{91.75}& 30.55&30.55&94.45&89.62 & 21.85& 21.90&94.20&88.24 &41.79&41.79 & 93.89&90.10\\
&DAG-FDD\cite{Ju2024improving}&18.64&25.33&94.35&87.19&30.92&30.92&\underline{96.23}&\underline{91.99} &19.45&19.45&94.91&91.49&30.44&30.44&\underline{95.22}&\textbf{90.89}\\
&PG-FDD\cite{lin2024preserving}&20.12&20.12&96.31&90.52&22.09&22.99&91.85&82.31&\underline{13.15}&\underline{15.62}&93.22&83.93&\underline{21.59}&\underline{24.05}&93.47&85.10\\ 
&\baseline{Ours}&\baseline{\textbf{10.60}}&\baseline{\underline{16.39}}&\baseline{\textbf{97.89}}&\baseline{91.57}&\baseline{\textbf{9.09}}&\baseline{\textbf{11.58}}&\baseline{\textbf{97.96}}&\baseline{\textbf{92.51}}&\baseline{\textbf{9.93}}&\baseline{\textbf{15.20}}&\baseline{\textbf{98.07}}&\baseline{\underline{91.93}}&\baseline{\textbf{19.87}}&\baseline{\textbf{19.87}}&\baseline{\textbf{97.70}}&\baseline{\underline{90.82}} \\ \hline

\multirow{5}{*}{Celebdf}
&Ori~\cite{Rossler_Cozzolino_Verdoliva_Riess_Thies_Niessner_2019}&29.52&\underline{29.52}&77.82&79.98&\underline{15.57}&\underline{15.57}&83.60&83.90&10.79&18.15&81.87&\textbf{84.33}&14.46&14.46&77.62&81.58\\
&DAW-FDD~\cite{Ju2024improving}&30.09&30.09&\underline{82.72}&\textbf{83.72}&36.12&26.12&\underline{84.81}&\underline{84.71} & 10.83 & \underline{15.85}& 76.20&66.57&\textbf{9.72}&\textbf{9.72}&78.78&81.85\\
&DAG-FDD\cite{Ju2024improving} &31.47&31.47&81.39&77.06&29.21&29.21&82.73&83.72&\underline{10.67}&\textbf{10.67}&83.19&82.82&30.30&30.30&\underline{83.39}&\textbf{84.15}\\
&PG-FDD~\cite{lin2024preserving}&\underline{27.37}&\textbf{27.37}&82.34&\underline{80.12}&20.86&20.86&80.56&82.06&21.35&24.28&\underline{84.21}&82.01&\underline{14.16}&\underline{14.16}&75.41&79.03\\ 
&\baseline{Ours}&\baseline{\textbf{21.02}}&\baseline{37.08}&\baseline{\textbf{85.11}}&\baseline{77.62}&\baseline{\textbf{12.45}}&\baseline{\textbf{12.45}}&\baseline{\textbf{89.25}}&\baseline{\textbf{86.36}}&\baseline{\textbf{7.50}}&\baseline{27.74}&\baseline{\textbf{88.08}}&\baseline{\underline{83.21}}&\baseline{16.84}&\baseline{17.89}&\baseline{\textbf{86.02}}&\baseline{\underline{80.67}} \\ \hline

\multirow{5}{*}{DFD}
&Ori~\cite{Rossler_Cozzolino_Verdoliva_Riess_Thies_Niessner_2019}&27.75&27.75&77.91&73.95&34.05&34.05&78.04&73.11&39.96&39.96&77.99&72.34&37.90&37.90&78.80&71.51\\
&DAW-FDD\cite{Ju2024improving} &\underline{18.38}&\underline{18.38}&78.39&73.09& \textbf{10.27}&35.45&72.67&68.59 & 21.74&21.74& 76.58& 68.44 &25.23&\textbf{25.23}&79.98&71.72\\
&DAG-FDD\cite{Ju2024improving} &25.15&25.15&80.56&72.53& 29.22 &\underline{29.22}& \underline{82.73}&\underline{73.72}&\underline{13.89}&\underline{13.89}&\underline{83.02}&\underline{73.96}&26.61&\underline{26.61}&81.09&\underline{73.91}\\
&PG-FDD~\cite{lin2024preserving}&\textbf{15.73}&\textbf{15.73}&\underline{81.53}&\underline{74.23}&30.76&30.76&74.46&67.91&16.93&34.58&78.33&70.43&\textbf{19.49}&27.90&\underline{81.09}&73.52\\ 

&\baseline{Ours}&\baseline{21.55}&\baseline{21.55}&\baseline{\textbf{86.70}}&\baseline{\textbf{79.64}}&\baseline{\underline{19.43}}&\baseline{\textbf{19.43}}&\baseline{\textbf{85.64}}&\baseline{\textbf{78.83}}&\baseline{\textbf{7.76}}&\baseline{\textbf{10.47}}&\baseline{\textbf{83.92}}&\baseline{\textbf{77.19}}&\baseline{\underline{20.57}}&\baseline{26.726}&\baseline{\textbf{84.87}}&\baseline{\textbf{78.05}} \\ \hline

\multirow{5}{*}{DFDC}
&Ori~\cite{Rossler_Cozzolino_Verdoliva_Riess_Thies_Niessner_2019}&15.61&49.45&62.07&\underline{59.86}&\underline{17.56}&\underline{42.41}&61.89&57.86&24.96&48.34&\underline{64.22}&\underline{61.88}&24.14&36.26&62.60&58.44\\
&DAW-FDD\cite{Ju2024improving} &\underline{13.99} &41.47& \textbf{64.15}&55.06& 27.03 & 45.56& 55.87& 53.21 &\textbf{22.13} & 49.71& 61.71& 60.56 &\underline{17.87}&\underline{33.79}&64.15&57.33\\
&DAG-FDD\cite{Ju2024improving}&26.26 &34.15& 56.84&51.87&20.82 & 55.53&\underline{64.02}&57.17&29.98&\textbf{27.63}&60.49&51.49&18.16& 38.38&\underline{64.16}&54.07\\
&PG-FDD~\cite{lin2024preserving}&22.36&\underline{31.92}&62.12&58.32&21.81&49.44&60.95&\underline{59.85}&\underline{22.82}&\underline{41.02}&57.33&60.28&20.74&56.67&58.75&\underline{59.78}\\ 

&\baseline{Ours}&\baseline{\textbf{10.91}}&\baseline{\textbf{29.92}}&\baseline{\underline{62.13}}&\baseline{\textbf{63.74}}&\baseline{\textbf{16.51}}&\baseline{\textbf{34.99}}&\baseline{\textbf{64.28}}&\baseline{\textbf{63.06}}&\baseline{24.45}&\baseline{48.80}&\baseline{\textbf{64.95}}&\baseline{\textbf{64.72}}&\baseline{\textbf{
7.79}}&\baseline{\textbf{21.46}}&\baseline{\textbf{65.14}}&\baseline{\textbf{65.17}} \\ \hline
% \multirow{5}{*}{Avg}
% &Ori~\cite{Rossler_Cozzolino_Verdoliva_Riess_Thies_Niessner_2019}&15.61&49.45&62.07&59.86&13.636&39.708&63.91&60.77&24.958&48.339&64.22&61.88&19.212&41.914&63.05&58.60\\
% &DAW-FDD\cite{Ju2024improving} &\underline{13.99} &41.47& \textbf{64.15}&55.06& 27.032 & 45.556& 55.87& 53.21 & 21.127 & 49.714& 67.71& 60.56 &17.873&33.790&64.15&57.33\\
% &DAG-FDD\cite{Ju2024improving}&26.26 &\underline{34.15}& 56.84&51.87&14.327 & 53.779&64.41&60.81 & 29.150& 32.130& 61.71&55.33&18.164& 38.383&64.16&54.07\\
% &PG-FDD~\cite{lin2024preserving}&16.20&35.30&\underline{62.90}&\textbf{65.66}&19.652&\textbf{27.778}&57.03&57.86&19.638&36.325&61.61&\textbf{61.02}&14.750&31.098&\textbf{63.69}&\textbf{63.92}\\ 
% &\baseline{Ours}&\baseline{\textbf{10.91}}&\baseline{\textbf{29.92}}&\baseline{62.13}&\baseline{\underline{63.74}}&16.511&34.990&64.28&63.06&24.454&48.795&64.95&64。72 \\ \hline
\end{tabular}
}
% \vspace{-3mm}
\label{tab:comparison_with_different_backbone}
\end{table*}

The previous work demonstrates splendid generalization performance on the intra-domain dataset. However, significant challenges remain in achieving fairness. The primary reason is that redundant semantics negatively impact the fairness performance of detectors.

As results in Table~\ref{tab:intra-domain}, our approach can refine the sensitivity of detection models to forgery artifacts, reducing the risk of over-parameterization caused by over-fitting to a particular forgery sub-dataset and redundant semantics. Compared with existing fairness approaches, the results demonstrate that our method improved detection performance while preserving demographic fairness.

\subsection{Evaluation for Fair Detection Performance on the Cross-domain Dataset}
To assess the fair performance and generalizability of our method in the cross-domain dataset. Table \ref{tab:cross-domain} shows the comparison results.
% All baselines are briefly described in the appendix \ref{appendix: Comparison Methods Overview}. 
All models are trained on FF++\_c23\cite{Rossler_Cozzolino_Verdoliva_Riess_Thies_Niessner_2019} and evaluated on four datasets under the same training and testing conditions. To simplify the comparison of fairness performance across different methods, we used $F_{MAG}$ as the fairness metric, evaluating the comprehensive detection performance on both positive and negative samples across various demographic subgroups within different cross-domain datasets.

The results indicate that fairness-enhanced detectors exhibit superior detection fairness, although the $AUC$ metric does not achieve the best performance. Additionally, frequency type detectors demonstrate better fairness performance compared to spatial type detectors, suggesting that frequency enhancement may help filter out certain demographic semantics irrelevant to forgery. Furthermore, our approach, in comparison to other fairness-enhanced detectors, achieves higher detection performance while maintaining superior fairness.

\begin{figure*}[t]
  \centering
  \includegraphics[width = 1\textwidth]{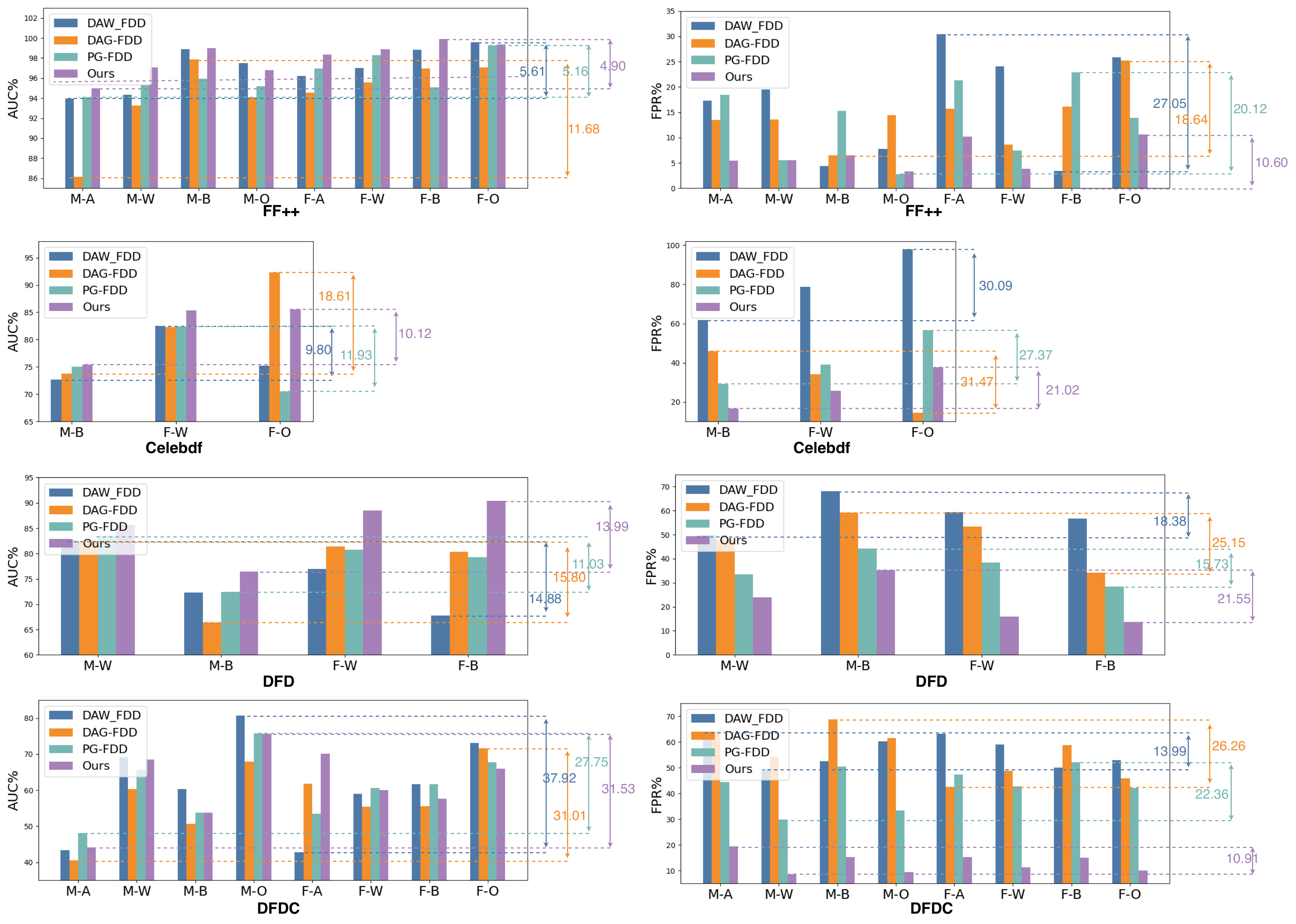}
  \caption{Detection performance of ours and three other fairness methods across different demographic subgroups within four datasets.}
  \label{fig:performance_subgroup}
  % \vspace{-10pt}
\end{figure*}

\begin{figure*}[tb]
  \centering
  \includegraphics[width = 1\textwidth]{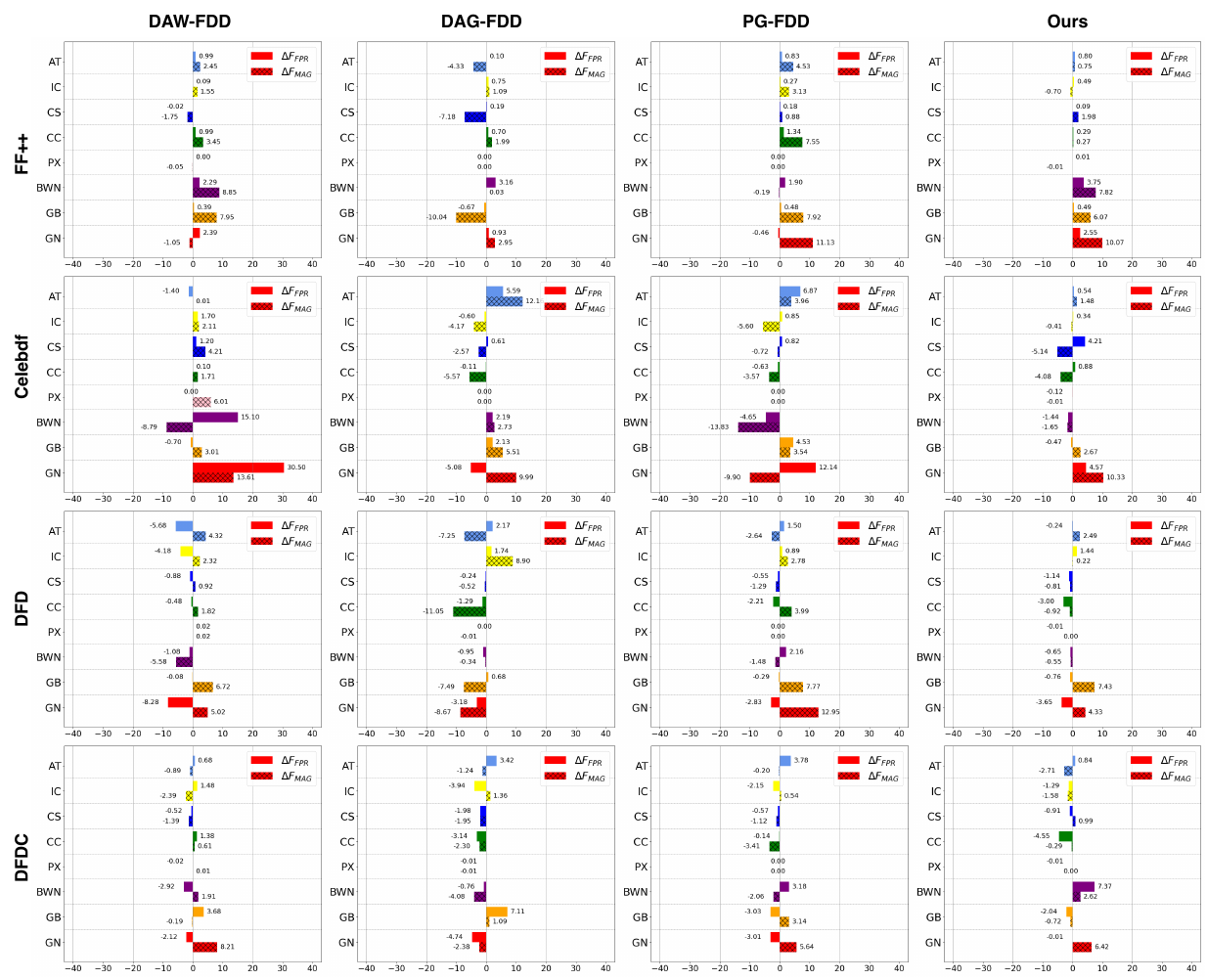}
  \caption{Evaluations for robustness of the proposed method. We examine the robustness of the proposed method against eight disturbances of various intensities. All models were trained and tested on the original FF++.}
  \label{fig:robustness}
\end{figure*}

\begin{table*}[th]
\centering
\caption{\centering{Fairness evaluation on different face pre-process.}}
\scalebox{0.72}{
\begin{tabular}{c|cccc|cccc|cccc|cccc}
\hline
\multirow{3}{*}{Pre-process} &
\multicolumn{4}{c|}{FF++}& \multicolumn{4}{c|}{Celebdf}&
\multicolumn{4}{c|}{DFD}& \multicolumn{4}{c}{DFDC}\\ \cline{2-17}
% srm 9
% dct 12
% 
&
\multicolumn{3}{c|}{\begin{tabular}[c]{@{}c@{}}Fairness\\ Metrics(\%)\textdownarrow\end{tabular}} & \multicolumn{1}{l|}{\begin{tabular}[c]{@{}c@{}}Detection\\ Metrics(\%)\textuparrow\end{tabular}} &
\multicolumn{3}{c|}{\begin{tabular}[c]{@{}c@{}}Fairness\\ Metrics(\%)\textdownarrow\end{tabular}} & \multicolumn{1}{l|}{\begin{tabular}[c]{@{}c@{}}Detection\\ Metrics(\%)\textuparrow\end{tabular}} &
\multicolumn{3}{c|}{\begin{tabular}[c]{@{}c@{}}Fairness\\ Metrics(\%)\textdownarrow\end{tabular}} & \multicolumn{1}{l|}{\begin{tabular}[c]{@{}c@{}}Detection\\ Metrics(\%)\textuparrow\end{tabular}} &
\multicolumn{3}{c|}{\begin{tabular}[c]{@{}c@{}}Fairness\\ Metrics(\%)\textdownarrow\end{tabular}} & \multicolumn{1}{l}{\begin{tabular}[c]{@{}c@{}}Detection\\ Metrics(\%)\textuparrow\end{tabular}}\\ \cline{2-17}

&
$F_{FPR}$&$F_{MAG}$&\multicolumn{1}{c|}{$F_{MEO}$}& $AUC$ & 
$F_{FPR}$&$F_{MAG}$& \multicolumn{1}{c|}{$F_{MEO}$} & $AUC$ & 
$F_{FPR}$ &$F_{MAG}$& \multicolumn{1}{c|}{$F_{MEO}$} & $AUC$ &
$F_{FPR}$ &$F_{MAG}$& \multicolumn{1}{c|}{$F_{MEO}$} & $AUC$ \\ \hline

DCT&\underline{19.08}&\underline{4.06}&\underline{19.08}&\textbf{98.21}&\textbf{20.33}&18.95&\textbf{26.34}&\underline{83.12}&42.01&15.72&42.01&\underline{83.75}&\textbf{8.49}&\textbf{27.27}&49.88&61.05\\ 
SRM&19.64&5.32&19.64&97.46&34.29&\underline{17.55}&34.29&83.06&\textbf{20.31}&\textbf{6.69}&\textbf{20.31}&80.52&24.50&\underline{30.68}&\textbf{27.47}&\underline{63.10}\\ 
w/o pre-process&22.50&\textbf{4.05}&22.50&97.85&30.12&35.35&\underline{30.12}&79.81&34.82&\underline{7.85}&34.82&82.50&22.39&36.56&48.89&\textbf{63.41}\\ \hline

\baseline{\begin{tabular}[c]{@{}c@{}}Ours\\ (astray-srm)\end{tabular}}&\baseline{\textbf{10.60}}&\baseline{4.90}&\baseline{\textbf{16.39}}&\baseline{\underline{97.89}}&\baseline{\underline{21.02}}&\baseline{\textbf{10.12}}&\baseline{37.08}&\baseline{\textbf{85.11}}&\baseline{\underline{21.55}}&\baseline{13.99}&\baseline{\underline{21.55}}&\baseline{\textbf{86.70}}&\baseline{10.91}&\baseline{31.53}&\baseline{\underline{29.92}}&\baseline{62.13}\\ \hline
\end{tabular}
}
\label{tab:comparison_with_different_preprocess}
\end{table*}

\begin{figure*}[t]
  \centering
  \includegraphics[width = 1\textwidth]{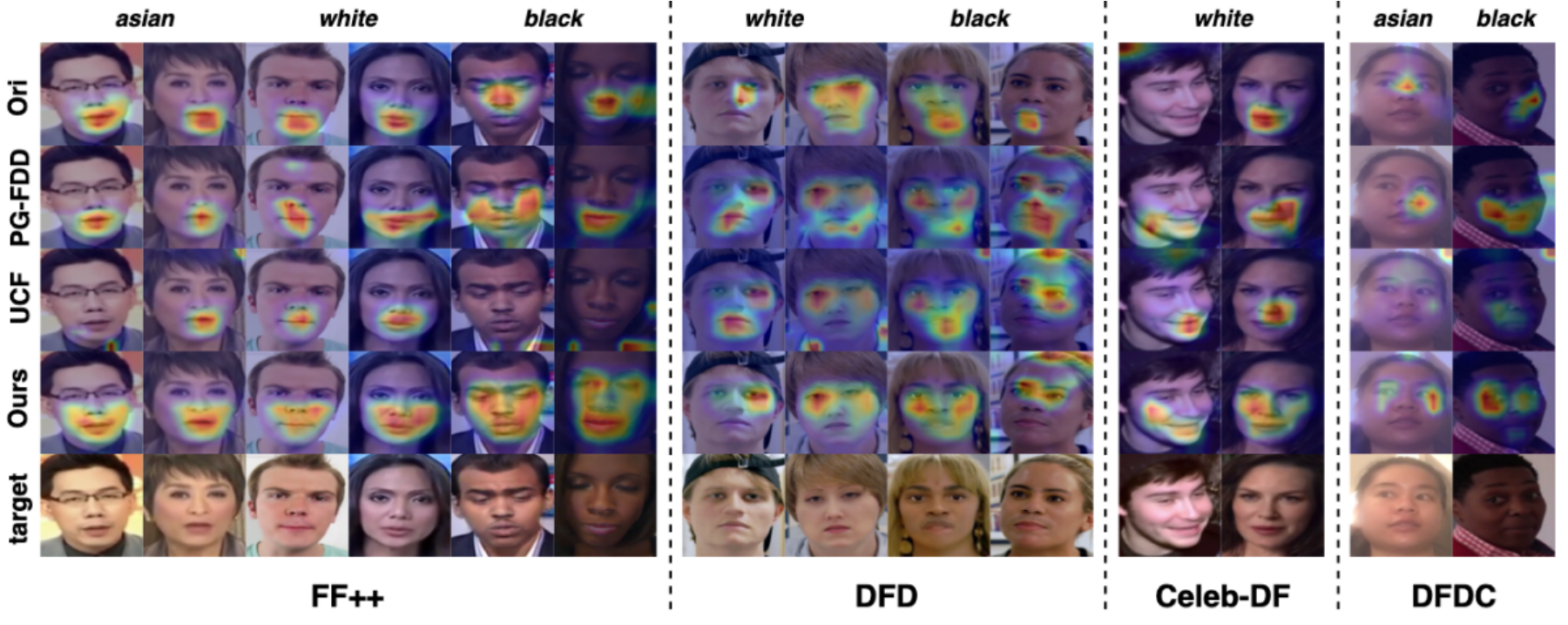}
  \caption{Grad-CAM visualization of ours and other three methods across four datasets with face images from different ethnic groups.}
  \label{fig:grad_cam}
  % \vspace{-10pt}
\end{figure*}

\begin{figure}[t]
  \centering
  \includegraphics[width = 0.48\textwidth]{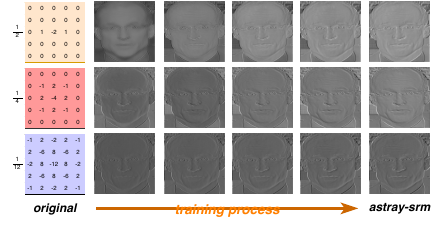}
  % \caption{Astray-SRM's adaptive change process in Misleading-learning.}
  \caption{The high-frequency image of Deepfakes processed by Astray-SRM is shown, with redundant content semantics effectively eliminated as the adaptive update processing of Astray-SRM continuously evolves.}
  \label{fig:astray-srm}
  % \vspace{-10pt}
\end{figure}

\subsection{Evaluation for Fair Generalizability of Different Backbones} 
To evaluate the fair generalizability based on different backbones.
We selected several classic backbones pre-trained on FF++: Xception \cite{Chollet_2017}, ResNet-50 \cite{He_Zhang_Ren_Sun_2016}, ResNet-32 \cite{He_Zhang_Ren_Sun_2016}, EfficientNet-B4 \cite{tan2019efficientnet}. Furthermore, to evaluate the fairness performance differences across fairness-enhanced detectors, we introduce two additional fairness metrics: $F_{FPR}$ and $F_{MEO}$. The results in Table \ref{tab:comparison_with_different_backbone} demonstrate the superiority of the proposed method for forgery semantics refinement. Note that, in our approach, we only adjusted the structure of the adapter, while the \textit{Misleading Separation} phase consistently employs the Xception as the backbone.

\begin{table*}[tb]
\centering
\caption{\centering{Ablation study for the loss functions and the fusion block of Misleading-learning. Specifically, `$\mathcal{L}_{cls}$' and `$\mathcal{L}_{con}$' represent our classification loss and the contrastive loss, respectively. `$\mathcal{B}_{scam}$' denotes single channel attention fusion block.}}
\scalebox{1.1}{
\begin{tabular}{cccc|cccccccc}
\hline
\multicolumn{4}{c|}{} & \multicolumn{8}{c}{Dataset} \\ \cline{5-12} 
\multicolumn{4}{c|}{\multirow{-2}{*}{Method}} & \multicolumn{2}{c|}{FF++} & \multicolumn{2}{c|}{Celebdf} & \multicolumn{2}{c|}{DFD} & \multicolumn{2}{c}{DFDC} \\ \cline{1-12} 

\multicolumn{1}{c|}{Name} & $\mathcal{D}_{bias}$ &$\mathcal{L}_{con}$&$\mathcal{B}_{scam}$&  $F_{FPR}$\textdownarrow & \multicolumn{1}{c|}{AUC\textuparrow} & $F_{FPR}$\textdownarrow & \multicolumn{1}{c|}{AUC\textuparrow} & $F_{FPR}$\textdownarrow & \multicolumn{1}{c|}{AUC\textuparrow} & $F_{FPR}$\textdownarrow & AUC\textuparrow \\ \hline

\multicolumn{1}{c|}{VariantA} && \checkmark  & \multicolumn{1}{c|}{\checkmark} &20.10& \multicolumn{1}{c|}{97.37} & 20.48& \multicolumn{1}{c|}{83.74} & 32.68& \multicolumn{1}{c|}{75.84} & 14.30& 61.41\\
\multicolumn{1}{c|}{VariantB} & \checkmark & &\multicolumn{1}{c|}{\checkmark} & 12.58 & \multicolumn{1}{c|}{98.40} & 20.83& \multicolumn{1}{c|}{83.17} & 41.72 & \multicolumn{1}{c|}{85.62} &11.72 & 62.36  \\ 
\multicolumn{1}{c|}{VariantC} & \checkmark & \checkmark& \multicolumn{1}{c|}{}  & 11.84& \multicolumn{1}{c|}{98.12} & 25.93 & \multicolumn{1}{c|}{84.52} & 42.93 & \multicolumn{1}{c|}{80.30} &10.40 & 63.01\\ \hline

\multicolumn{1}{c|}{\baseline{Ours}} &\baseline{\checkmark} & \baseline{\checkmark} & \multicolumn{1}{c|}{\baseline{\checkmark}} & \baseline{10.60} & \multicolumn{1}{c|}{\baseline{97.89}} & \baseline{21.02} & \multicolumn{1}{c|}{\baseline{85.11}} & \baseline{21.55} & \multicolumn{1}{c|}{\baseline{86.70}} & \baseline{10.91} & \baseline{62.13} \\ \hline
\end{tabular}
}

\label{tab:ablation}
\end{table*}

\subsection{Evaluation for Detection Performance for Each Demographic Subgroup}
To more effectively quantify the fair performance disparities between our method and other fairness-enhanced approaches, we computed the fairness performance of four fairness methods across all subgroups. We employed two metrics: AUC and FPR, and illustrated the results in Fig~\ref{fig:performance_subgroup}. The left column of the figure illustrates the $AUC$ metric across different demographic subgroups, while the right column displays the $FPR$ metric for the same subgroups. The double-arrow vertical lines to the right of each bar chart indicate the maximum difference in detection metrics across different demographic subgroups. Shorter vertical lines suggest better fairness. For the $AUC$ metric, the higher the midpoint of the double-arrow vertical line, the better the overall $AUC$ metric. For the $FPR$ metric, the lower the midpoint of the double-arrow vertical line, the better the overall $FPR$ metric. It is evident that our method, in comparison to other fairness approaches, narrows the detection performance gap among different demographic subgroups while sustaining a higher average detection accuracy, thereby ensuring superior fairness performance.

\subsection{Evaluation for Robustness}

In addition to evaluating the generalization and fairness of a model, it's crucial to test its robustness, as there might be inevitable information loss due to post-processing. A more robust model can adapt more effectively to detection in the real world.
Following the previous works \cite{jiang2020deeperforensics,haliassos2021lips}, we implemented 8 types of video disturbances.
% , each with 5 intensity levels.

Fig~\ref{fig:robustness} reports the fairness and detection performance differences for each method under different disturbance types compared to the no-interference scenario. We introduce two additional metrics: $\Delta F_{FPR}$ and $\Delta F_{MAG}$. The corresponding manipulations for GN, GB, BWN, PX, CC, CS, IC, and AT are Gaussian Noise, Gaussian Blur, Block-Wise Noise, Pixelation, Color Contrast, Color Saturation, Image Compression, and Affine Transformation, respectively.
% Below is the meaning of each abbreviation, GN: Gaussian Noise, GB: Gaussian Blur, BWN: Block-Wise Noise, PX: Pixelation, CC: Color Contrast, CS: Color Saturation, IC: Image Compression, AT: Affine Transformation.

Observed from the experimental results, our method exhibits great robustness in most cases. Despite the disentanglement-based method being more robust on Gaussian blur and Gaussian noise, the proposed method manifests greater robustness against Block-Wise noise and Affine Transformations.

\subsection{Evaluation for Fair Generalizability of Different Image Pre-processes}
\label{sec:Evaluation for Fair Generalizability of Different Image Pre-processes.}

We used DCT and SRM to preprocess the image for Misleading-learning, respectively.
Table~\ref{tab:comparison_with_different_preprocess} shows the comparison results. All methods were trained on FF++ and evaluated on four datasets. The results indicate that Misleading-learning with DCT preprocessing leads to significant performance degradation within DFDC. Additionally, utilizing the original image (w/o pre-process) for Misleading-learning increases the complexity of the training process. When compared to the original \textit{SRM}, the proposed \textit{Astray-SRM} exhibited a significant boost in performance.
Furthermore, we present the filtered results after the dynamic updates of \textit{Astray-SRM}, as shown in Fig~\ref{fig:astray-srm}.

\subsection{Visualization}

% \paragraph{Visualization of the Saliency Map. }

\label{Visualization of the Saliency Map}
 The Grad-CAM\cite{selvaraju2017grad} provides a clear visualization demonstrating the effectiveness of our method. We visualize \cite{Yan2023UCFUC, lin2024preserving, Rossler_Cozzolino_Verdoliva_Riess_Thies_Niessner_2019} and the proposed method in Fig~\ref{fig:grad_cam}. Visualizations from various data sets and facial attributes reveal that the baseline tends to overfit small, localized regions due to its lack of specialized generalization enhancement. PG-FDD and UCF are two methods that aim to improve generalization and fairness through a disentanglement framework. While they succeed in reducing over-fitting to local regions, they remain susceptible to noise from areas outside the face. 
 In contrast, our method, whether intra-domain or cross-domain, demonstrates greater resistance to noise originating from outside the face area, with an increased focus on the significant facial features.

\subsection{Ablation Study}
During the overall training phase of our approach, performance is significantly impacted by three key factors, the design of the selection bias for pristine face, Misleading-learning loss function, and the feature fusion block. Therefore, we conduct ablation experiments on all components, including the selection bias, the contrastive loss function, and the feature fusion block.
The contribution of each component is shown in the results of Variant A/B/C in Table~\ref{tab:ablation}. Variant A/B/C emphasizes the selection bias, the contrastive loss, and the feature fusion block value in the Misleading-learning process, respectively( e.g., within the FF++ dataset, in the absence of $\mathcal{D}_{bias}$, $F_{FPR}$ increases $9.50\%$ and $AUC$ drops $0.59\%$).

\section{Conclusion}
We propose misleading-learning which features refining facial forgery semantics in the paper. The exhaustive evaluations manifest that the proposed method could mitigate the negative impact of detrimental and irrelevant semantics, thereby enhancing the fairness, as well as generalizability, and robustness for DeepFake detectors. Also, the effectiveness of the proposed modules is justified by ablation studies. Unlike the traditional approaches of employing end-to-end large models to confront AI-generated content counterfeiting, the proposed method provides the technical framework and foundation for developing more powerful methods for detecting falsified faces as well as other AI-generated content in the future.

% \paragraph{Limitation}

% Although our approach considers a broader spectrum of unknown detrimental semantics, it cannot achieve higher fairness over the disentanglement framework designed for specific semantics. The inherent unfairness of the dataset itself should also be taken into account.

{\small
\bibliographystyle{IEEEtran}
\bibliography{Astray-Learning}
}

\vfill
\clearpage

\end{document}